%% file: bare_jrnl_new_sample4.tex
\documentclass[lettersize,journal]{IEEEtran}
\usepackage[OT1]{fontenc}
\usepackage{amsmath,amsfonts}
\usepackage{algorithm}
\usepackage{array}
\usepackage{graphicx}
\usepackage{cite}
\usepackage{algpseudocode}
\usepackage{tabularx}
\usepackage{float}     
\usepackage{placeins}
\usepackage{subcaption}
\usepackage{booktabs}

\usepackage[normalem]{ulem}
\usepackage{xcolor}

\newcommand{\best}[1]{\textbf{#1}}
\newcommand{\secondbest}[1]{\uline{#1}}
\newcommand{\thirdbest}[1]{\emph{#1}}

\newif\ifshowrevisions
\showrevisionsfalse
\ifshowrevisions
    
    \newcommand{\revdel}[1]{\textcolor{red}{\sout{#1}}}
    \newcommand{\revise}[2]{\revdel{#1}\revadd{#2}}
    \newcommand{\revnote}[1]{\textcolor{magenta}{[Revision note: #1]}}
\else
    
    \newcommand{\revdel}[1]{}
    \newcommand{\revise}[2]{#2}
    \newcommand{\revnote}[1]{}
\fi

\begin{document}

\title{3DGAA: Realistic and Robust 3D Gaussian-based Adversarial Attack\\for Autonomous Driving}

\author{
Yixun Zhang, Lizhi Wang, Junjun Zhao, Wending Zhao, Feng Zhou, Yonghao Dang, and Jianqin Yin*\thanks{*Corresponding author.} \\
School of Intelligent Engineering and Automation,
Beijing University of Posts and Telecommunications, China \\
{\tt\small \{zhangyixun, wanglizhi, zhaojunjun, windyz, zhoufeng, dyh2018, jqyin\}@bupt.edu.cn}
}


\maketitle

\input{sec/00_abstract}   
\input{sec/01_intro}
\input{sec/02_relatedc}

\input{sec/03_methodc}
\input{sec/04_experimentsc}
% \input{sec/05_discussion}
\input{sec/10_conclusion}
% \clearpage

% \section{Conclusion}
% The conclusion goes here.

\section*{Acknowledgments}
% This work was supported partly by the National Natural Science Foundation of China (Grant No. 62173045, 62273054), partly by the Fundamental Research Funds for the Central Universities (Grant No. 2020XD-A04-3), and the Natural Science Foundation of Hainan Province (Grant No. 622RC675).

This work was supported partly by the Beijing Natural Science Foundation (F2024203115), partly by the State Grid Corporation Science and Technology Project Funding (5700-202422243A-1-1-ZN), partly by the Taishan Industrial Experts Program, and the National Natural Science Foundation of China (62403491).

% {\appendix[Proof of the Zonklar Equations]
% Use $\backslash${\tt{appendix}} if you have a single appendix:
% Do not use $\backslash${\tt{section}} anymore after $\backslash${\tt{appendix}}, only $\backslash${\tt{section*}}.
% If you have multiple appendixes use $\backslash${\tt{appendices}} then use $\backslash${\tt{section}} to start each appendix.
% You must declare a $\backslash${\tt{section}} before using any $\backslash${\tt{subsection}} or using $\backslash${\tt{label}} ($\backslash${\tt{appendices}} by itself
%  starts a section numbered zero.)}


{
\appendix
\input{sec/12_appendix}

}

% \section{References Section}
% You can use a bibliography generated by BibTeX as a .bbl file.
%  BibTeX documentation can be easily obtained at:
%  http://mirror.ctan.org/biblio/bibtex/contrib/doc/
%  The IEEEtran BibTeX style support page is:
%  http://www.michaelshell.org/tex/ieeetran/bibtex/
 
%  % argument is your BibTeX string definitions and bibliography database(s)
% %\bibliography{IEEEabrv,../bib/paper}
% You can manually copy in the resultant .bbl file and set second argument of $\backslash${\tt{begin}} to the number of references
%  (used to reserve space for the reference number labels box).

% \bibliographystyle{IEEEtran}

{
    \small
    \bibliographystyle{IEEEtran}
    \bibliography{sec/11_references}
}

% \end{thebibliography}

\newpage

\section{Biography Section}
 

% \end{IEEEbiography}



\vspace{-33pt}\begin{IEEEbiography}[{\includegraphics[width=1in,height=1.25in,clip,keepaspectratio]{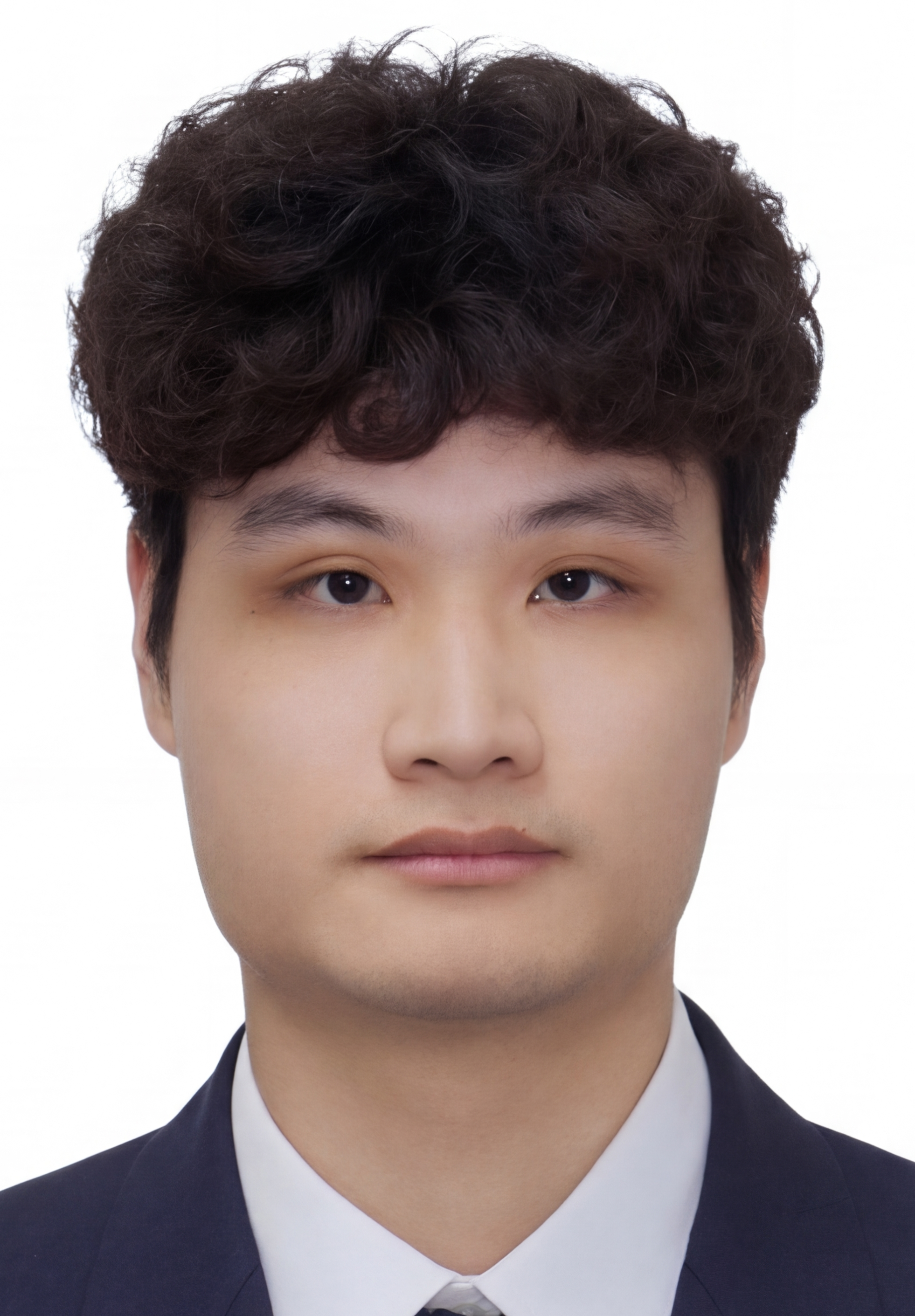}}]{Yixun Zhang}
    received his bachelor's degree from Xi'an University of Posts and Telecommunications, Shaanxi, China, in 2023, and is currently a master's degree candidate at the School of Intelligent Engineering and Automation, Beijing University of Posts and Telecommunications, Beijing, China. His research interests include computer vision, adversarial attacks and AI safety.
\end{IEEEbiography}
\vspace{-33pt}\begin{IEEEbiography}[{\includegraphics[width=1in,height=1.25in,clip,keepaspectratio]{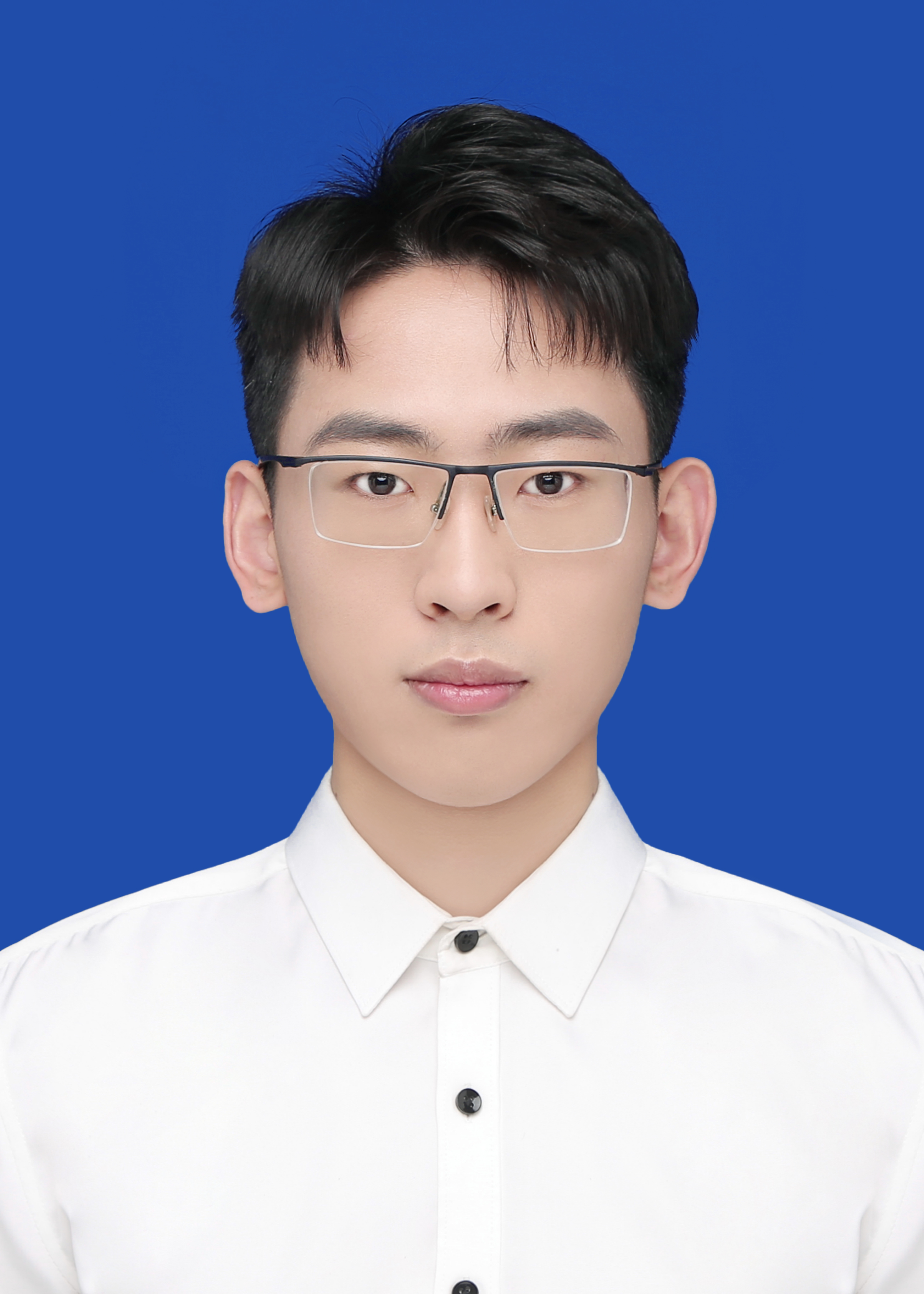}}]{Lizhi Wang}
    received his bachelor's degree from Beijing University of Posts and Telecommunications, Beijing, China, in 2023, and is currently a master's degree candidate at the School of Intelligent Engineering and Automation, Beijing University of Posts and Telecommunications. His research interests include computer vision, machine learning and 3D vision.
\end{IEEEbiography}
\vspace{-33pt}\begin{IEEEbiography}[{\includegraphics[width=1in,height=1.25in,clip,keepaspectratio]{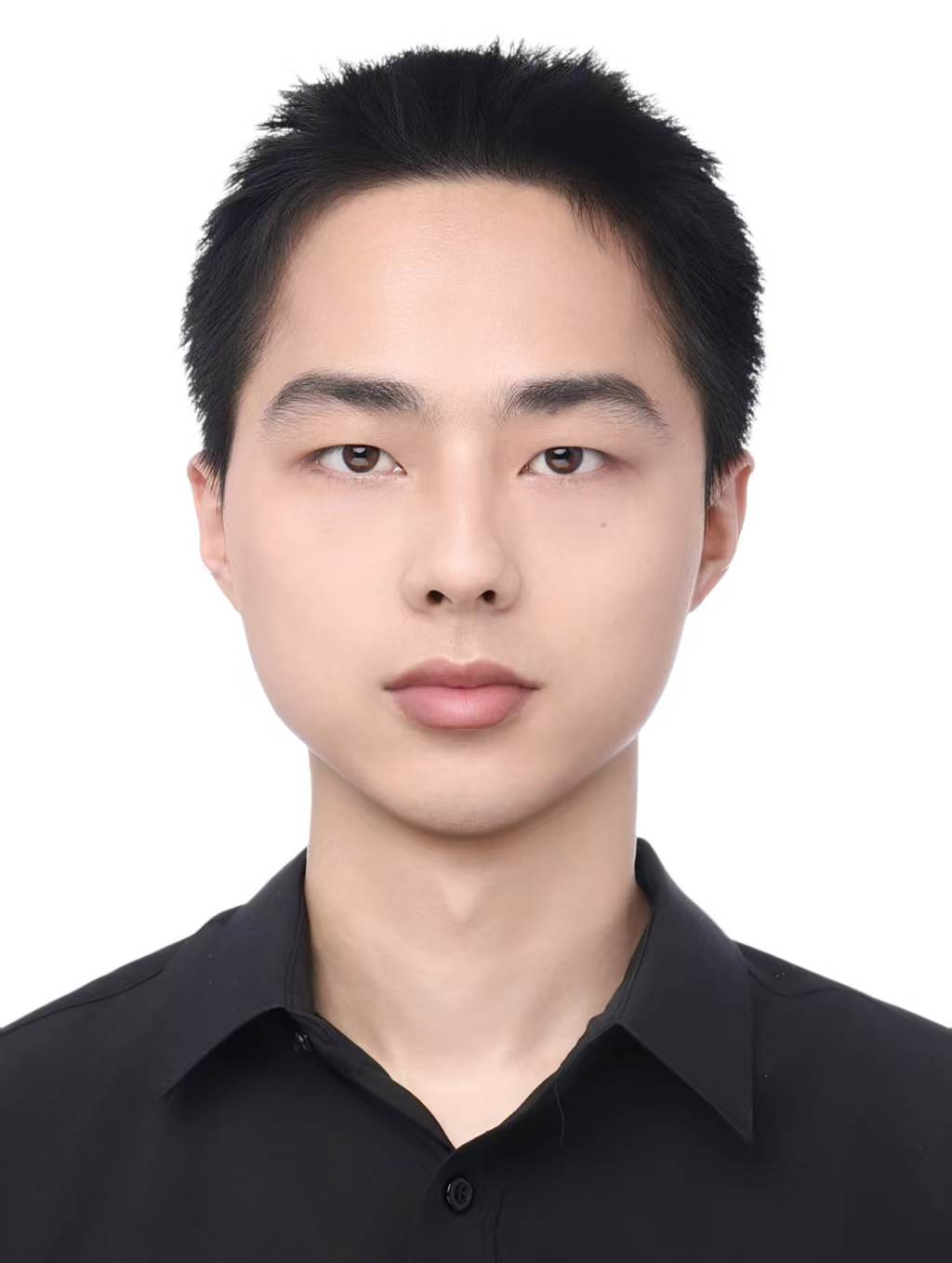}}]{Junjun Zhao}
    received his bachelor's degree from Beijing University of Posts and Telecommunications, Beijing, China, in 2023, and is currently a master's degree candidate at the School of Intelligent Engineering and Automation, Beijing University of Posts and Telecommunications. His research interests include computer vision, image processing and 3D virtual humans.
\end{IEEEbiography}
\vspace{-33pt}\begin{IEEEbiography}[{\includegraphics[width=1in,height=1.25in,clip,keepaspectratio]{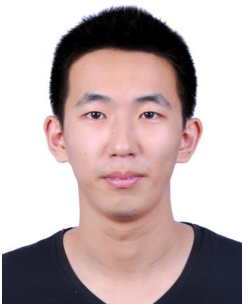}}]{Wending Zhao}
    received his bachelor's degree from Beijing University of Posts and Telecommunications, Beijing, China, in 2023, and is currently a master's degree candidate at the School of Intelligent Engineering and Automation, Beijing University of Posts and Telecommunications. His research interests include connected and autonomous vehicles, traffic networks and AI safety.
\end{IEEEbiography}
\vspace{-33pt}\begin{IEEEbiography}[{\includegraphics[width=1in,height=1.25in,clip,keepaspectratio]{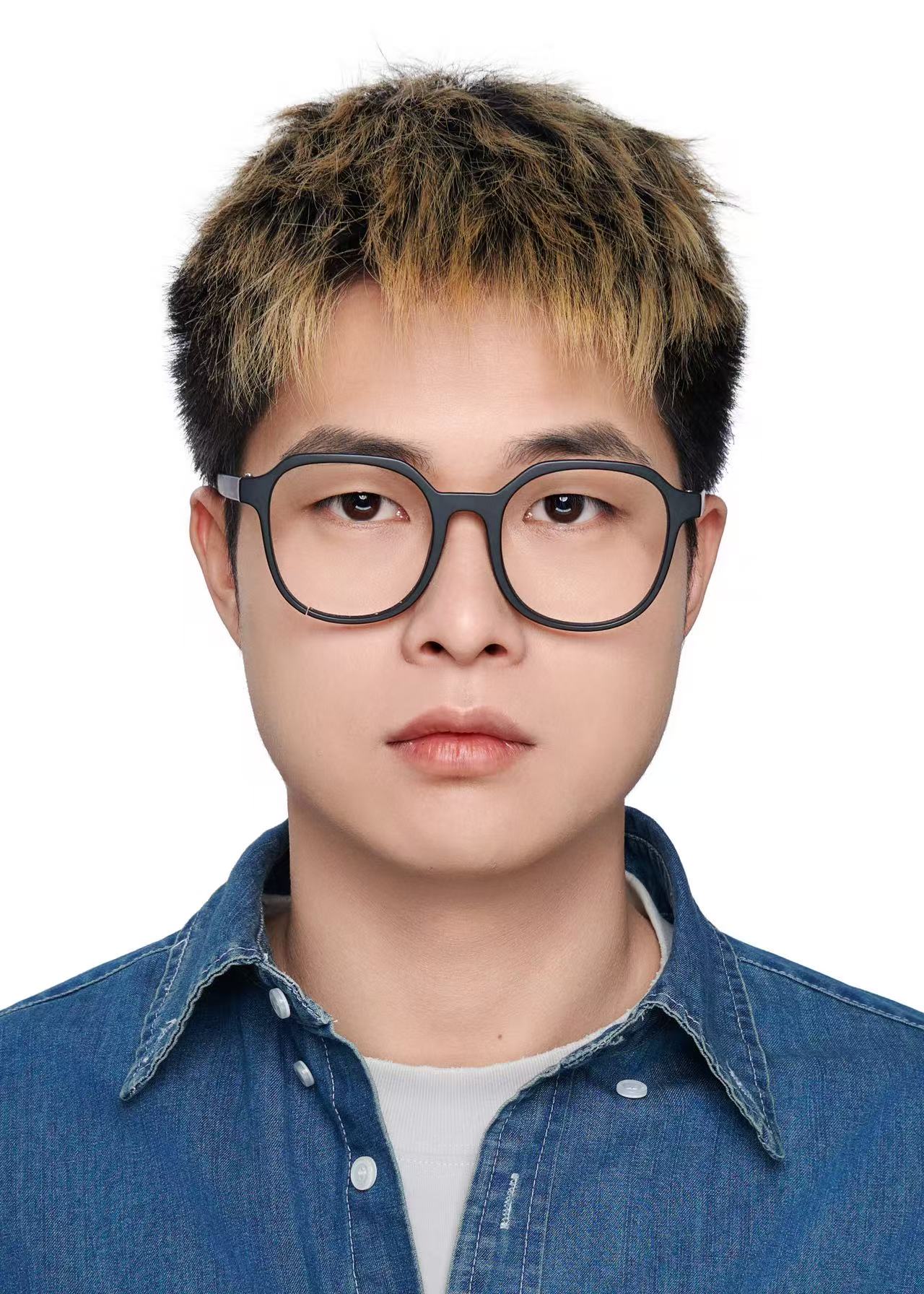}}]{Feng Zhou}
    received his bachelor's degree from Beijing University of Posts and Telecommunications, Beijing, China, in 2022, and is currently a Ph.D. candidate at the School of Intelligent Engineering and Automation, Beijing University of Posts and Telecommunications. His research interests include computer vision, visual generative models and 3D virtual humans.
\end{IEEEbiography}
\vspace{-33pt}\begin{IEEEbiography}[{\includegraphics[width=1in,height=1.25in,clip,keepaspectratio]{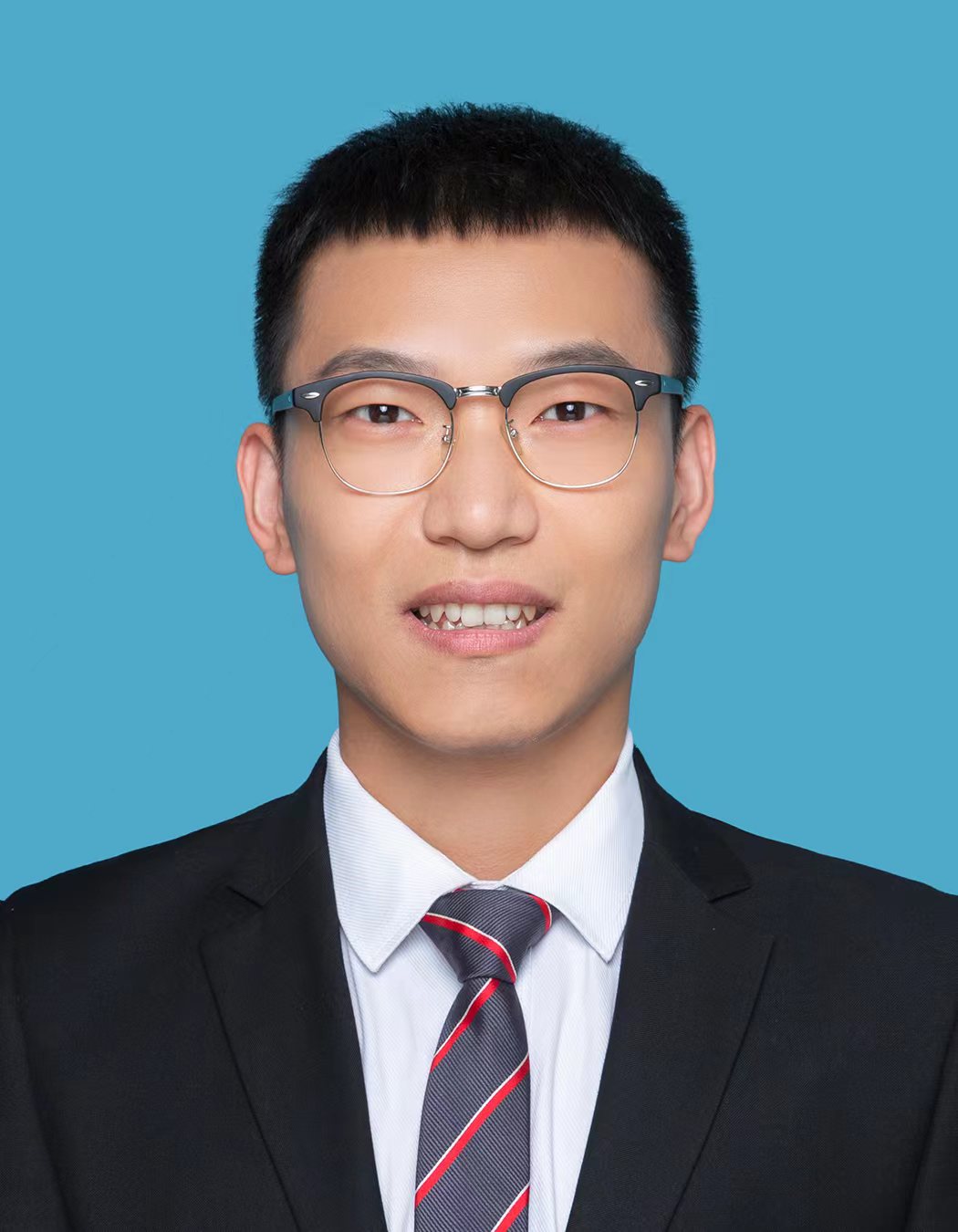}}]{Yonghao Dang}
    received his bachelor's degree in computer science and technology from the University of Jinan, Jinan, China, in 2018. He is currently a Ph.D. candidate with the School of Intelligent Engineering and Automation, Beijing University of Posts and Telecommunications, Beijing, China. His research interests include computer vision, image processing, and deep learning.
\end{IEEEbiography}
\vspace{-33pt}\begin{IEEEbiography}[{\includegraphics[width=1in,height=1.25in,clip,keepaspectratio]{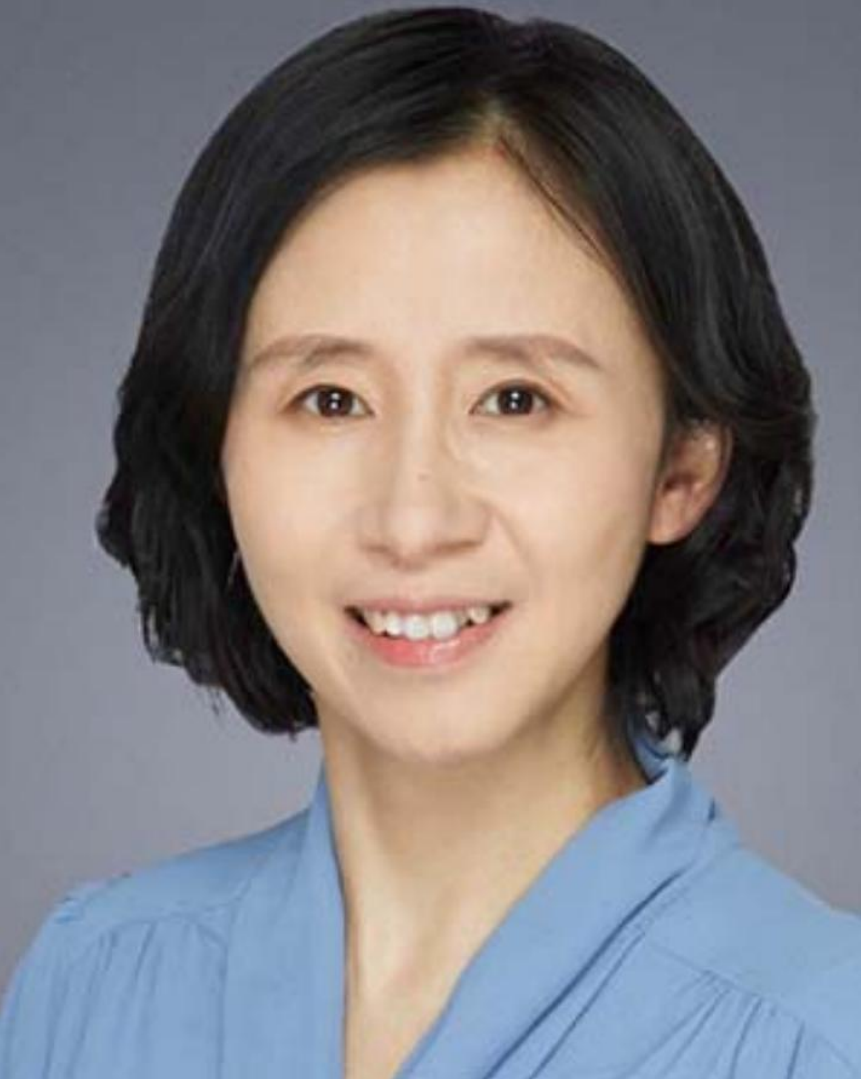}}]{Jianqin Yin}
    received the Ph.D. degree from Shandong University, Jinan, China, in 2013. She currently is a Professor with the School of Intelligent Engineering and Automation, Beijing University of Posts and Telecommunications, Beijing, China. Her research interests include service robot, pattern recognition, machine learning and image processing. Email: jqyin@bupt.edu.cn.
\end{IEEEbiography}

\vfill

\end{document}

%% file: sec/00_abstract.tex
\begin{abstract}

% Camera-based object detection systems play a vital role in autonomous driving, yet they remain vulnerable to adversarial threats in real-world environments. Existing 2D and 3D physical attacks, due to their focus on texture optimization, often struggle to balance physical realism and attack robustness. In this work, we propose \textbf{3D Gaussian-based Adversarial Attack (3DGAA)}, a novel adversarial object generation framework that leverages the full 14-dimensional parameterization of 3D Gaussian Splatting (3DGS) to jointly optimize geometry and appearance in physically realizable ways.
% Unlike prior works that rely on patches or texture optimization, 3DGAA jointly perturbs both geometric attributes (shape, scale, rotation) and appearance attributes (color, opacity) to produce physically realistic and transferable adversarial objects. We further introduce a \textbf{physical filtering module} that filters outliers to preserve geometric fidelity, and a \textbf{physical augmentation module} that simulates complex physical scenarios to enhance attack generalization under real-world conditions.
% We evaluate 3DGAA on both virtual benchmarks and physical-world setups using miniature vehicle models. Experimental results show that 3DGAA achieves to reduce the detection mAP from \textbf{87.21\%} to \textbf{7.38\%}, significantly outperforming existing 3D physical attacks. Moreover, our method maintains high transferability across different physical conditions, demonstrating a new state-of-the-art in physically realizable adversarial attacks. 

Camera-based perception in connected and autonomous vehicles remains exposed to physical adversarial attacks. 
Prior attacks often either optimize image-plane textures, weakening cross-view consistency, or rely on shape modifications that are difficult to fabricate and deploy. Both cases limit their value for security and safety assessment of camera-based perception.
This paper introduces a fabrication-first framework that learns view-consistent, geometry-preserving adversarial wraps for vehicles. The method performs three-dimensional multi-view optimization with a 3D Gaussian splatting surrogate to enforce consistency under an expectation over viewpoint, illumination, and occlusion, while the final artifact is constrained to print-only textures that keep vehicle geometry unchanged. 
We evaluate the framework through matched CARLA simulations, controlled miniature-vehicle physical evidence, and tests on multiple modern detectors. The wraps produce substantial and stable reductions in detection confidence and average precision across views while maintaining perceptual realism. Ablation studies isolate the effects of physical filtering, physical augmentation, and shape-consistency regularization, while efficiency analyses report convergence time and memory profiles. Additional experiments report ASR, mIoU-based segmentation degradation, cross-detector transfer against physical attack baselines, perturbation-area fairness diagnostics, and robustness under common input preprocessing defenses. 
By coupling three-dimensional optimization with a geometry-preserving realization, the study supports a practical pathway to systematically stress-test camera perception in safety-critical autonomous driving systems and provides evidence that can inform defense design and evaluation protocols in information forensics and security.

\end{abstract}

\begin{IEEEkeywords}
Adversarial machine learning; Information forensics; Physical adversarial attacks; Computer vision; Autonomous driving; 3D Gaussian splatting.
\end{IEEEkeywords}

%% file: sec/01_intro.tex
\section{Introduction}
\label{sec:intro}

Camera-based object detection systems are a backbone of connected and autonomous vehicles (CAVs)~\cite{autoware,baidu,tesla}, forming a security-critical component for core functions such as obstacle avoidance and lane keeping. However, a growing body of work shows that these models remain vulnerable to physical adversarial artifacts. In the physical world, subtle and structured perturbations can induce missed detections or misclassifications~\cite{6248074,Carlini_Wagner_2017, Goodfellow_Shlens_Szegedy_2014,ren_adversarial_2020, multiview, Surveillance}, which raises concerns for safety and security assurance in downstream planning for safety-critical scenarios. Systematically exploring such vulnerabilities is therefore essential to understand and improve the real-world robustness of perception modules in autonomous driving~\cite{Brown_Mane_Roy_Abadi_Gilmer_2017, Chen_Cornelius_Martin_Chau_2018, Thys_Ranst_Goedeme_2019,Zhang_Foroosh_David_Gong_2018, Huang_Gao_Zhou_Xie_Yuille_Zou_Liu_2020, Wu_Ning_Li_Huang_Yang_Wang_2020, Wang_Liu_Yin_Liu_Tang_Liu_2021, Wang_Jiang_Sun_Zhou_Gong_Zhang_Yao_Chen_2022, Suryanto_Kim_Kang_Larasati_Yun_Le_Yang_Oh_Kim, suryanto2023activehighlytransferable3d,huang2024towards, lou20253dgaussiansplattingdriven, RAUCA, PACG}.

Existing physical adversarial attacks can be understood through three related but insufficient design choices: appearance perturbation, shape perturbation, and transformation-based robustness training. 2D methods attach adversarial patches to objects~\cite{Brown_Mane_Roy_Abadi_Gilmer_2017, Chen_Cornelius_Martin_Chau_2018, Thys_Ranst_Goedeme_2019, Huang_Gao_Zhou_Xie_Yuille_Zou_Liu_2020}, which preserves physical realism through minimal appearance changes but often exhibits limited robustness across viewpoints or environmental conditions (\textbf{Challenge 1}). In contrast, recent 3D attacks~\cite{Zhang_Foroosh_David_Gong_2018, Wu_Ning_Li_Huang_Yang_Wang_2020, Wang_Liu_Yin_Liu_Tang_Liu_2021, Wang_Jiang_Sun_Zhou_Gong_Zhang_Yao_Chen_2022, Suryanto_Kim_Kang_Larasati_Yun_Le_Yang_Oh_Kim, suryanto2023activehighlytransferable3d, lou20253dgaussiansplattingdriven} manipulate surface textures to achieve stronger and more consistent multi-view performance. Yet many such methods remain primarily texture-driven: they change colors or camouflage patterns but do not jointly control the geometry and appearance cues that detectors use across views, which can introduce visual artifacts or unrealistic surface distortions and restrict applicability in safety-critical deployment (\textbf{Challenge 2}). Shape- or geometry-based attacks offer additional degrees of freedom, but unconstrained body deformation is difficult to fabricate on real vehicles and can violate the non-deformability constraints of practical wrap-based deployment. Standard expectation-over-transformation (EoT) improves robustness by averaging over sampled views or photometric changes, but it does not by itself specify a fabricable 3D object representation, enforce cross-view consistency, or couple the optimization with physical filtering constraints. These observations reveal an inherent trade-off between \emph{physical realism} and \emph{adversarial robustness}, creating a key obstacle for real-world use where both goals are required (\textbf{Challenge 3}). 

To address these challenges, we introduce \textbf{3DGAA}, a fabrication-first adversarial object generation framework that uses 3D optimization to learn view-consistent, physically constrained vehicle wraps. The framework leverages \emph{3D Gaussian Splatting} (3DGS)~\cite{kerbl3Dgaussians}, a differentiable and compact 3D representation originally designed for photorealistic rendering. The key role of 3DGS is not merely to replace a mesh or texture map. It provides a shared object-level parameterization in which position, scale, rotation, opacity, and color are optimized through the same differentiable renderer, so an adversarial update from one view must remain compatible with the same object when rendered from other views. Unlike mesh- or point-based representations, 3DGS encodes shape and texture in a unified 14-dimensional parameter space, which enables fine-grained and physically consistent adversarial manipulation~\cite{hong2024lrmlargereconstructionmodel,tang2024dreamgaussiangenerativegaussiansplatting,tang2024lgmlargemultiviewgaussian}. For deployment-oriented evaluation in safety-critical perception systems, we adopt a fabrication-first, geometry-preserving realization that produces print-only wraps, while the full joint optimization provides an analytical upper-bound setting in controlled evaluation.

3DGAA comprises three modules that jointly reconcile realism and robustness. 
First, a \emph{Physically-Constrained Adversarial Optimization} stage optimizes the 14-dimensional Gaussian parameters (position, scale, rotation, opacity, color) under geometry-preserving and box constraints, and minimizes a detector-aware loss aggregated over an expectation of transforms in viewpoint and distance as well as over multiple detectors. A shape-consistency regularization limits normal with respect to the source object. This procedure suppresses detector confidence consistently across views while complying with printable, geometry-preserving realization, addressing \textbf{Challenge 1}. 
Second, a \emph{Physical Filtering Module} applies topology pruning to remove sparsely supported or isolated splats and performs structural denoising on scale and opacity fields while preserving silhouettes. These operations convert the vanilla 3DGS reconstruction into a filtered 3DGS that serves as the initial object for adversarial optimization, eliminating ringing and fold-over artifacts typical of texture-only attacks and improving multi-view visual plausibility, thereby tackling \textbf{Challenge 2}. 
Third, a \emph{Physical Augmentation Module} injects stochastic, physically motivated variations--including photometric shifts, sensor noise, cast shadows, occlusions, and mild pose jitter--directly into the optimization loop through expectation-over-transform sampling. Different from using EoT as a standalone image-space robustness wrapper, our augmentation is tied to the same 3DGS object, the same multi-view rendering distribution, and the same geometry-preserving fabrication constraints; therefore, the optimized adversarial signal must survive both environmental variation and object-level view consistency. By optimizing risk over this augmented distribution, the learned wraps retain their effect under tested illumination and viewpoint changes, which improves generalization in deployment-like conditions and addresses \textbf{Challenge 3}.

We evaluate 3DGAA under a matched experimental protocol that combines virtual benchmarks with controlled miniature-vehicle physical evidence. Specifically, 3DGAA reduces detection mAP from \textbf{87.21\%} to \textbf{7.38\%} and achieves \textbf{91.39\%} ASR in simulation, while the physical study verifies a geometry-preserving print-and-apply path under representative lighting conditions. As shown in Fig.~\ref{fig:intro}, 3DGAA occupies a favorable region in the realism--robustness space relative to representative 2D and 3D baselines. We further examine transformer-based detectors, semantic segmentation, cross-detector transfer against physical attack baselines, input-preprocessing defenses, and perturbation-area fairness, thereby clarifying both effectiveness and evaluation boundaries. Our contributions are summarized as follows:

\begin{itemize}
    \item We propose \textbf{3DGAA}, a fabrication-first 3DGS-based adversarial framework for camera-based perception in autonomous driving. It moves beyond texture-only attacks by optimizing geometry and appearance in a unified 14-dimensional Gaussian parameter space, while retaining a geometry-preserving, print-only realization path for physical deployment.
    \item We formulate a physically constrained optimization pipeline that uses a single differentiable 3D object to enforce multi-view consistency. The full model jointly optimizes geometric and appearance attributes to characterize adversarial capacity, whereas the deployment variant freezes geometry and transfers the optimized appearance to printable wraps.
    \item We design two modules that enhance practicality and robustness in safety- and security-critical perception scenarios: a \textbf{physical filtering module} based on topological pruning and structural denoising to enforce geometric fidelity, and a \textbf{physical augmentation module} that couples EoT-style environmental sampling with 3DGS rendering and fabrication constraints to improve robustness under viewpoint, illumination, sensor, and occlusion changes.
    \item We present extensive evaluation in simulation and with controlled physical miniature vehicles. The study covers detection mAP and ASR, mIoU-based segmentation transfer, transformer-based detectors, cross-detector transfer against physical attack baselines, input-preprocessing defenses, perturbation-area fairness, and condition-level physical evidence, which together provide a practical pathway to stress-test camera perception in safety-critical autonomous driving systems.
\end{itemize}

%% file: sec/02_relatedc.tex
\begin{figure}[t]
    \centering
    \begin{subfigure}[b]{0.48\linewidth}
        \includegraphics[width=\linewidth]{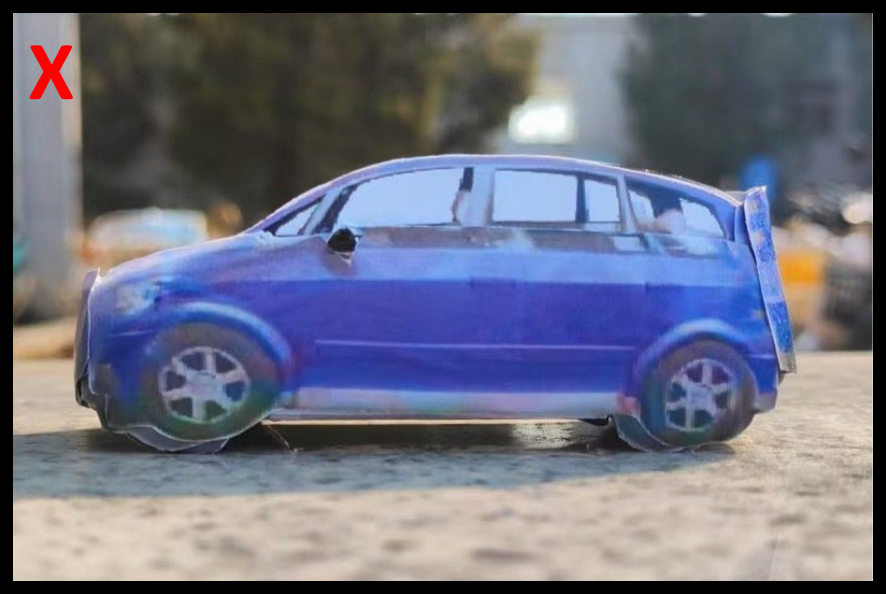}
        \caption{Controlled physical setup: the wrapped object is missed by the detector.}
    \end{subfigure}
    \hfill
    \begin{subfigure}[b]{0.48\linewidth}
        \includegraphics[width=\linewidth]{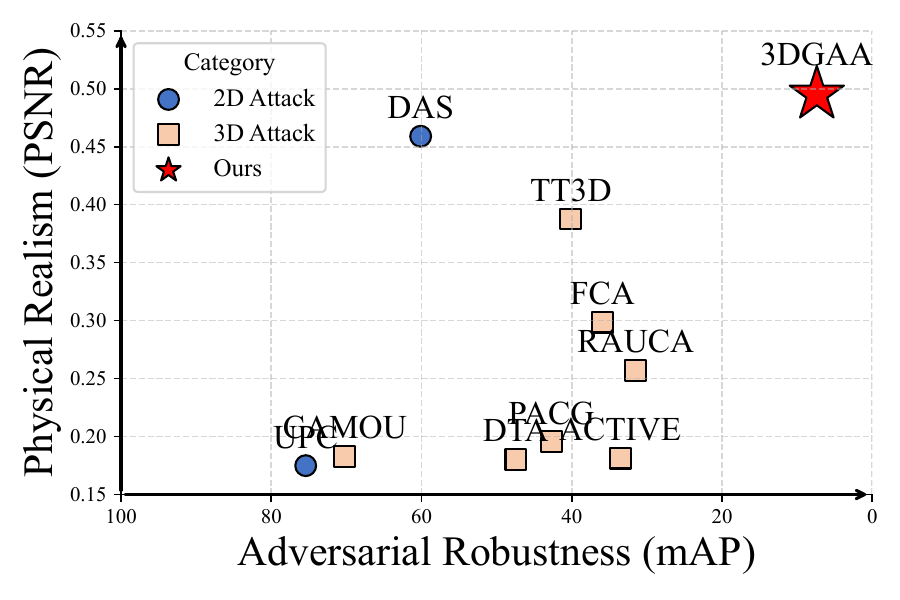}
        \caption{Realism vs. Robustness comparison across methods.}
    \end{subfigure}
    \vspace{-2mm}
    \caption{
    (a) Controlled physical setup of 3DGAA on a miniature vehicle, resulting in detection failure under the tested condition.
    (b) Comparison of adversarial methods in terms of physical realism (PSNR) and attack effectiveness (mAP). 3DGAA occupies a favorable realism--robustness region among the evaluated baselines.
    }
    \label{fig:intro}\vspace{-2mm}
\end{figure}

\section{Related Work}
\label{sec:related}

\textbf{Physical adversarial attacks.} In autonomous driving scenarios, physical adversarial attacks aim to deceive real-world perception systems by perturbing object appearance in a deployable manner. Early methods~\cite{Brown_Mane_Roy_Abadi_Gilmer_2017, Chen_Cornelius_Martin_Chau_2018, Thys_Ranst_Goedeme_2019, liu2018dpatch, lee2019physical} mainly adopt 2D patch-based perturbations rendered on printable surfaces. Robust traffic-sign attacks further show that physical adversarial examples must account for changing scale, blur, resolution, illumination, and video-time capture conditions in autonomous-vehicle perception~\cite{jia2022foolingeyes}. UPC~\cite{Huang_Gao_Zhou_Xie_Yuille_Zou_Liu_2020} introduces universal physical patches that generalize across scenarios, DAS~\cite{Wang_Liu_Yin_Liu_Tang_Liu_2021} exploits differentiable mesh rendering~\cite{Kato_Ushiku_Harada_2018} to inject perturbation via attention manipulation. Although these approaches offer strong attack performance, they often exhibit limited robustness across viewpoints and ambient illumination typical of CAV operation.

To improve cross-view generalization, recent works leverage 3D object representations and render adversarial textures onto full object meshes~\cite{Zhang_Foroosh_David_Gong_2018, Wu_Ning_Li_Huang_Yang_Wang_2020, Wang_Liu_Yin_Liu_Tang_Liu_2021, Wang_Jiang_Sun_Zhou_Gong_Zhang_Yao_Chen_2022, Suryanto_Kim_Kang_Larasati_Yun_Le_Yang_Oh_Kim, RAUCA, PACG}. In particular, FCA~\cite{Wang_Jiang_Sun_Zhou_Gong_Zhang_Yao_Chen_2022} generates full-surface camouflage with physical constraints. These approaches~\cite{Suryanto_Kim_Kang_Larasati_Yun_Le_Yang_Oh_Kim, suryanto2023activehighlytransferable3d}, however, primarily operate in the appearance space and rarely couple appearance with geometric cues. Recent works such as TT3D~\cite{huang2024towards} and PGA~\cite{lou20253dgaussiansplattingdriven} explore high-dimensional texture search spaces.

Despite their improved transferability, most existing physical attacks rely solely on texture perturbations, which limits expressiveness and yields brittleness under mild geometric misalignment and sensor/illumination shifts encountered in CAV operation~\cite{hull2024adversarialattacksusingdifferentiable, 10.1145/3469877.3493596, 10.1145/3636551}. These methods lack the ability to jointly manipulate geometric cues such as shape, scale, or position, factors critical for consistent real-world perception. Conversely, methods that substantially deform object shape can increase attack freedom but are difficult to realize on rigid vehicle bodies without violating practical manufacturing and deployment constraints. Moreover, fabrication constraints and geometry-preserving realizations are rarely explicitly modeled, which hinders applicability in regulated transportation settings.
EoT-style training~\cite{athalye2018synthesizingrobustadversarialexamples} is also widely used to improve robustness under sampled transformations, but in many physical attacks it remains an image- or texture-space training strategy rather than a complete object-level formulation. It does not automatically provide a shared 3D parameter space, a filtering mechanism for geometric artifacts, or a path from optimized representation to a printable geometry-preserving artifact.

To overcome these limitations, we propose a fabrication-first physical attack framework that extends beyond appearance-only changes.

Our approach leverages 3D Gaussian Splatting~\cite{kerbl3Dgaussians}, a differentiable and expressive 3D representation, to jointly optimize both geometry and appearance in a unified 14-dimensional space~\cite{10521791,chen2025survey3dgaussiansplatting,tang2024lgmlargemultiviewgaussian,hong2024lrmlargereconstructionmodel,Li_2025_WACV}. This formulation enables physically realistic perturbations that remain effective under deployment-like viewpoint and imaging variations~\cite{NDSDF,zhang2024cameras}. Unlike prior works constrained to textures, our method introduces perturbation through joint optimization of geometry and appearance, enabling physically realistic and adversarially robust object perturbations for evaluating robust perception systems. The distinction is therefore methodological rather than only representational: 3DGAA combines object-level 3DGS optimization, physical filtering, EoT-style environmental sampling, and a fabrication-first deployment branch so that novelty, robustness, and physical feasibility are optimized together.

%% file: sec/03_methodc.tex
\section{Method}
\label{sec:method}

\begin{figure*}[t]
    \centering
    \includegraphics[width=\linewidth]{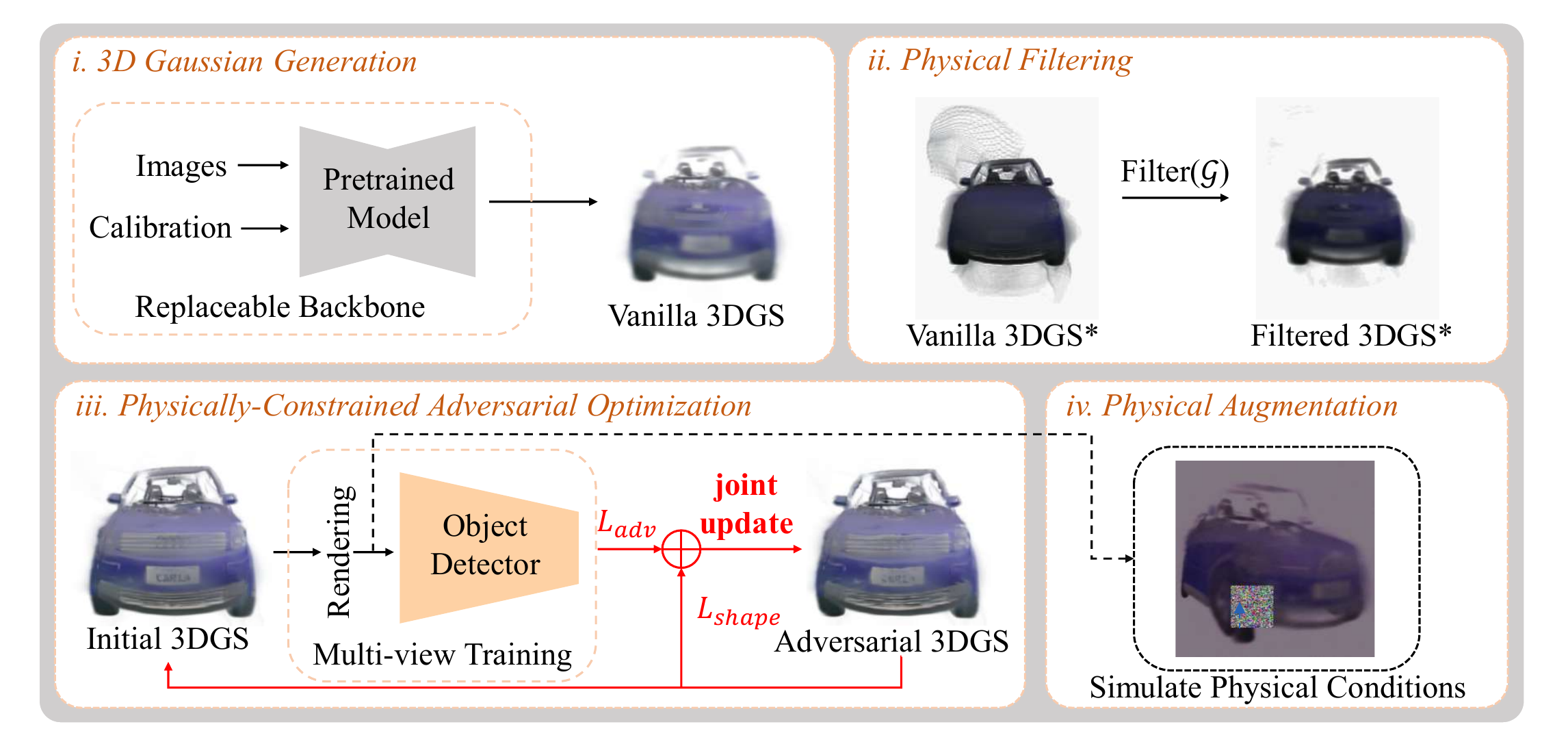}\vspace{-4mm}
    \parbox{\linewidth}{\raggedright \scriptsize \vspace{0.5mm}* indicate display-only visual processing, where brightness is reduced and contrast is enhanced to make the filtering effect easier to inspect. }
    \caption[Overview of the proposed 3DGAA pipeline.]{
    \textbf{Overview of the proposed 3DGAA pipeline.}
    (i) Given multi-view images and camera calibration, a pretrained backbone generates a vanilla 3D Gaussian Splatting (3DGS) object.
    (ii) The Physical Filtering Module applies topological pruning and structural denoising to the vanilla object, producing the filtered 3DGS used as the initial 3DGS for adversarial optimization. 
    (iii) The adversarial optimization stage directly updates the same 3DGS object with $L_{\text{adv}}$ and $L_{\text{shape}}$, yielding an adversarial 3DGS. Thus, the Initial 3DGS and Adversarial 3DGS denote the same underlying object before and after optimization, rather than two independently rendered images. This object-level gradient update distinguishes 3DGAA from image-space attacks that modify only rendered views.
    (iv) Physical augmentation simulates deployment-time conditions for the optimized adversarial object.
    }

    \label{fig:overall}\vspace{-2mm}
\end{figure*}

\subsection{Preliminaries}
\label{sec:preliminaries}

% Our goal is to generate physically realistic adversarial 3D objects that can degrade the performance of camera-based object detection models under real-world conditions.

Formally, we assume access to $n$ multi-view RGB images $\{I_i\}_{i=1}^n$ of a target object, captured under calibrated camera poses $\{P_i\}_{i=1}^n$. These inputs are fed into a pretrained 3D Gaussian Splatting (3DGS) generation network to generate a base object $\mathcal{G} = \{\mathbf{g}_j\}_{j=1}^N$, where each Gaussian primitive $\mathbf{g}_j \in \mathbb{R}^{14}$ is parameterized by: Position $\mathbf{x} \in \mathbb{R}^3$, Rotation $\mathbf{q} \in \mathbb{R}^4$ (quaternion), Scale $\mathbf{s} \in \mathbb{R}^3$, Color $\mathbf{c} \in \mathbb{R}^3$, Opacity $\alpha \in \mathbb{R}$.
This 14-dimensional representation allows for fine-grained control over both geometry and appearance, making it highly suitable for physical adversarial object generation. The 3DGS generation process can be formulated as:

\begin{equation}
    \mathcal{G} = \mathcal{F}_{\text{3DGS}}(\{I_i, P_i\}_{i=1}^n),
\end{equation}

\noindent where $\mathcal{F}_{\text{3DGS}}$ denotes the pretrained Gaussian generation network that maps calibrated RGB images to a set of $N$ Gaussians $\mathcal{G} = \{\mathbf{g}_j\}_{j=1}^N$, with each $\mathbf{g}_j \in \mathbb{R}^{14}$ encoding geometry and appearance attributes.

\noindent\textbf{Optimizing target}. Our goal is to perturb the Gaussian parameters such that the object, when rendered into a scene and processed by a detector $D$, minimizes the predicted confidence for its true class. This objective also defines the adversarial loss used throughout the optimization:
\begin{equation}
    \min_{\mathcal{G}} \; \mathcal{L}_{\text{adv}} =
    \mathbb{E}_{P \sim \mathcal{V}} \left[ f_D(R(\mathcal{G}, P)) \right].
    \label{equ:adv_objective}
\end{equation}

\noindent where $\mathcal{G}$ is the optimized 3DGS object, $P$ is a camera pose sampled from the viewpoint distribution $\mathcal{V}$, $R(\mathcal{G}, P)$ is the differentiable rendering of $\mathcal{G}$ from pose $P$, $D$ is the target detector, and $f_D(\cdot)$ returns the detector confidence for the target class. Minimizing $\mathcal{L}_{\text{adv}}$ therefore directly reduces the expected target-class confidence over multiple views.

\subsection{Overview}
\label{sec:overview}

As shown in Figure~\ref{fig:overall}, our proposed \textbf{3DGAA} framework generates adversarial 3D objects through a four-stage pipeline: 
\textbf{i. 3D Gaussian Generation:} Multi-view images and camera parameters are fed into a pretrained backbone to generate an initial 3DGS object. 
\textbf{ii. Physical Filtering:} Structural artifacts and topological noise are removed to enhance physical realism. 
\textbf{iii. Adversarial Optimization:} The filtered object is optimized using an adversarial loss $L_{\text{adv}}$ and a shape consistency loss $L_{\text{shape}}$. 
\textbf{iv. Physical Augmentation:} Differentiable augmentations simulate environmental conditions to improve robustness under deployment-like variations. 
Together, these modules produce adversarial objects that are both physically realistic and robust across the tested viewpoints and environmental variations. The mechanistic rationale for this design is summarized below before the detailed optimization procedure.

% As shown in Figure~\ref{fig:overall}, our proposed framework \textbf{3DGAA} generates adversarial 3D objects through a structured four-stage pipeline: \textbf{3D Gaussian Generation.} Multi-view images and camera parameters are used to generate a vanilla 3DGS representation via a pretrained backbone. \textbf{Physical Filtering.} The raw 3DGS object is denoised and structurally refined through topological pruning and artifact suppression, enhancing its geometric plausibility. \textbf{Physically-Constrained Adversarial Optimization.} The filtered 3DGS is adversarially optimized using two losses: (i) $L_{\text{adv}}$, which minimizes detector confidence; (ii) $L_{\text{shape}}$, which constrains geometry to remain physically realizable. \textbf{Physical Augmentation.} To improve robustness under real-world deployment, various physical effects are injected during training via differentiable augmentation.
% These components work together to produce adversarial objects that are both realistic and adversarially robust, enabling robust physical attacks across varying viewpoints and environmental conditions.

\subsection{Mechanistic Rationale}
\label{sec:mechanistic_rationale}

3DGAA derives its multi-view consistency from optimizing a single 3D Gaussian object rather than independent 2D views. Each primitive contributes to every rendered view according to the same center, covariance, opacity, and color, while the camera pose only changes the projection operator. Therefore, a parameter update that reduces detector confidence from one pose changes the underlying object that will be re-rendered from other poses. When the adversarial objective is optimized over a pose distribution $\mathcal{V}$, the gradients from multiple views are accumulated on the same Gaussian parameters. This shared-parameter mechanism discourages view-specific artifacts and encourages adversarial patterns that remain compatible across viewpoints.

The 14-dimensional Gaussian parameterization also explains why 3DGAA can jointly control geometry and appearance. The position, scale, and rotation parameters determine where each primitive is placed, how large its anisotropic support is, and how it contributes to the object's apparent shape. The color and opacity parameters determine the visible texture, intensity, and local transparency of the rendered object. Detector features are affected by both groups: geometric parameters influence contours, part locations, and projected scale, whereas appearance parameters influence local texture and color contrast. Optimizing these parameters in the same differentiable rendering loop allows 3DGAA to search coupled geometry-appearance perturbations, while the dimension-selection switch later separates appearance-only, geometry-only, and full modes when physical constraints require different realization paths.

Physical filtering improves deployability by removing adversarial changes that are effective numerically but implausible as physical objects. Direct high-dimensional optimization can create isolated Gaussians, abrupt scale changes, or unstable opacity spikes. These artifacts may lower detector confidence in rendered images but can disappear, flicker, or become visually implausible under different views. Topological pruning removes sparsely supported primitives, and structural denoising smooths scale and opacity fields in a camera-aware manner. The resulting representation keeps adversarial signal on coherent object surfaces, which supports fabrication and reduces multi-view rendering artifacts.

Physical augmentation improves robustness by optimizing the same 3D object under deployment-time nuisance factors. Unlike a post-processing test-time perturbation, the augmentation module is inside the training loop, so gradients are computed after photometric shifts, sensor noise, shadows, occlusions, and pose perturbations. The optimized parameters therefore minimize detector confidence in expectation over both viewpoint and environmental variation. This mechanism explains why the final appearance-only realization can retain its effect after printing and imaging, even though the physical vehicle geometry remains unchanged.

\subsection{Physically-Constrained Adversarial Optimization}
\label{sec:optimization}

To address the limited perturbation capability of traditional methods, we propose a physically-constrained adversarial optimization framework that perturbs both the geometry and texture of the 3DGS representation. This high-degree-of-freedom space enables more expressive and robust adversarial behavior under varying physical conditions.

% Given a generated 3D Gaussian object $\mathcal{G} = \{\mathbf{g}_j\}_{j=1}^N$, where each primitive $\mathbf{g}_j$ is defined by 14-dimensional parameters including position $\mathbf{p}_j$, scale $\mathbf{s}_j$, rotation $\mathbf{q}_j$, color $\mathbf{c}_j$, and opacity $\alpha_j$, our goal is to optimize $\mathcal{G}$ such that the rendered images fool a pretrained object detector $D$.

\vspace{1mm}
\noindent\textbf{Adversarial Loss.}
To suppress the detector's confidence on the target class, we use the adversarial objective in Eq.~\ref{equ:adv_objective} as $\mathcal{L}_{\text{adv}}$ throughout this stage, with all associated symbols defined at the first formulation.

\vspace{1mm}
\noindent\textbf{Shape Consistency Loss.}
To preserve the physical plausibility of the object's geometry, we introduce a shape constraint that penalizes deviations from the original structure:
\begin{equation}
\mathcal{L}_{\text{shape}} = \frac{1}{N} \sum_{j=1}^N \left\| \mathbf{p}_j - \mathbf{p}_{j,0} \right\|_2^2 + \left\| \mathbf{s}_j - \mathbf{s}_{j,0} \right\|_2^2 + \left\| \mathbf{q}_j \otimes \mathbf{q}_{j,0}^{-1} \right\|_2^2,
\end{equation}
where $\mathbf{p}_{j,0}$, $\mathbf{s}_{j,0}$ and $\mathbf{q}_{j,0}$ are the initial position, scale and rotation before optimization. $\otimes$ denotes quaternion multiplication.
This shape consistency loss discourages the attack from concentrating its effect in large geometric deformation, which would be difficult to realize as a physically plausible object.

% \vspace{1mm}
% \noindent\textbf{Total Loss and Optimization.}
% The final objective combines the adversarial and shape consistency terms:
% \begin{equation}
% \label{equ:total loss}
% \mathcal{L} = \lambda_{\text{adv}} \cdot \mathcal{L}_{\text{adv}} + \lambda_{\text{shape}} \cdot \mathcal{L}_{\text{shape}}.
% \end{equation}
% To dynamically balance attack strength and geometric fidelity, we set $\lambda_{\text{adv}}$ to be inversely proportional to the adversarial loss and $\lambda_{\text{shape}}$ to follow the unnormalized shape deviation. This allows the optimization to prioritize realism when the attack is already strong, and vice versa. 

% Algorithm~\ref{alg:adv-opt} summarizes the optimization loop. The 3DGS parameters are iteratively updated via gradient descent, where each iteration involves rendering, applying physical augmentations, computing adversarial and shape losses, and selectively applying gradients to relevant parameter subsets. The dynamic weights $\lambda_{\text{adv}}$ and $\lambda_{\text{shape}}$ are adjusted according to the magnitude of the current loss to balance the strength of the attack and the preservation of the realism. Further details are provided in Appendix~\ref{sup: dynamic loss weighting}.

\textbf{Total Loss and Optimization.}
The final objective combines the adversarial and shape consistency terms:
\begin{equation}
    \mathcal{L} = \lambda_{\mathrm{adv}} \cdot \mathcal{L}_{\mathrm{adv}} 
    + \lambda_{\mathrm{shape}} \cdot \mathcal{L}_{\mathrm{shape}}.
    \label{equ:total_loss}
\end{equation}

Instead of fixing $\lambda_{\mathrm{adv}}$ and $\lambda_{\mathrm{shape}}$ by hand, 
we adapt them online from the relative magnitudes of the two losses. 
Let $\bar{\mathcal{L}}_{\mathrm{adv}}$ and $\bar{\mathcal{L}}_{\mathrm{shape}}$ denote 
exponentially smoothed versions of $\mathcal{L}_{\mathrm{adv}}$ and 
$\mathcal{L}_{\mathrm{shape}}$ (updated at each iteration with decay factor 
$\eta$). We first rescale the adversarial loss to a comparable range:
\begin{equation}
    \hat{\mathcal{L}}_{\mathrm{adv}} = \frac{\bar{\mathcal{L}}_{\mathrm{adv}}}{\gamma},
\end{equation}
where $\gamma$ is a constant scale factor (set to $10.0$ in all experiments).

We then compute raw weights
\begin{equation}
    w_{\mathrm{adv}} = \frac{1}{\hat{\mathcal{L}}_{\mathrm{adv}} + \varepsilon},
    \qquad
    w_{\mathrm{shape}} = \bar{\mathcal{L}}_{\mathrm{shape}},
\end{equation}
and normalize them to obtain the loss coefficients
\begin{equation}
    \lambda_{\mathrm{adv}} = 
    \frac{w_{\mathrm{adv}}}{w_{\mathrm{adv}} + w_{\mathrm{shape}} + \varepsilon},
    \qquad
    \lambda_{\mathrm{shape}} = 1 - \lambda_{\mathrm{adv}}.
\end{equation}
To avoid collapsing into a single objective, we further clamp the shape
weight to a minimum value $\lambda_{\min}$:
\begin{equation}
    \lambda_{\mathrm{shape}} = \max(\lambda_{\mathrm{shape}}, \lambda_{\min}),
    \qquad
    \lambda_{\mathrm{adv}} = 1 - \lambda_{\mathrm{shape}},
\end{equation}
with $\lambda_{\min} = 0.4$ in our implementation. 
This adaptive scheme reduces the emphasis on the adversarial term once the attack 
has saturated and allocates more weight to geometry preservation, leading to
more stable training and better trade-offs between realism and robustness.

Algorithm~\ref{alg:adv-opt} summarizes the full optimization loop. At each 
iteration, we render the current 3DGS, apply the Physical Augmentation Module, 
compute $\mathcal{L}_{\mathrm{adv}}$ and $\mathcal{L}_{\mathrm{shape}}$, update
$(\lambda_{\mathrm{adv}}, \lambda_{\mathrm{shape}})$ using the above rules, and 
perform gradient descent on the selected parameter subsets. 

\subsection{Source-Detector Optimization and Transfer Evaluation}
\label{sec:source_transfer_protocol}

The adversarial objective is optimized with respect to a source detector $D_s$, but the optimized 3DGS object is not tied to the source detector at evaluation time. After optimization, the Gaussian parameters are fixed and the same object is rendered under the evaluation views before being processed by unseen detectors. No gradient, confidence score, or detector-specific post-processing from the unseen detector is used to refine the object. This separation is important because the source detector defines how the adversarial object is generated, whereas transfer evaluation tests whether the learned object-level perturbation remains harmful after the decision function changes.

This protocol follows naturally from the 3DGS formulation. Since the attack updates a shared 3D object rather than independent rendered images, the adversarial signal is distributed over geometry-aware and appearance-aware parameters that are reprojected across views. Optimizing the expectation over viewpoints and augmentation states therefore reduces the tendency to fit a single camera pose or a single detector activation pattern. The transferability experiments in Sec.~\ref{sec:exp} use this source-to-unseen setting to distinguish detector-specific attack strength from object-level adversarial behavior.

% The optimization is performed using gradient descent over the differentiable 3DGS parameters, which is detailed in Section~\ref{sup: physically-constrained adversarial optimization}.

% \begin{figure}
%     \centering
%     \begin{subfigure}[b]{0.4\linewidth}
%         \centering
%         \includegraphics[width=\textwidth]{figs/180_0.png}
%         \caption{Vinilla 3DGS}
%     \end{subfigure}
%     \begin{subfigure}[b]{0.4\linewidth}
%         \centering
%         \includegraphics[width=\textwidth]{figs/180_1.png}
%         \caption{TP}
%     \end{subfigure}
%     \vspace{0.5cm}
%     \begin{subfigure}[b]{0.4\linewidth}
%         \centering
%         \includegraphics[width=\textwidth]{figs/180_2.png}
%         \caption{SD}
%     \end{subfigure}
%     \begin{subfigure}[b]{0.4\linewidth}
%         \centering
%         \includegraphics[width=\textwidth]{figs/180_3.png}
%         \caption{TP+SD}
%     \end{subfigure}
%     \caption{Multi-step filtering results: (a) raw 3DGS after adversarial optimization, (b) after topological pruning, (c) after structural denoising, (d) combined filtering. Our method eliminates outlier primitives and smooths surface structure.}
%     \label{fig:filtering}
% \end{figure}

\subsection{Physical Filtering Module}
\label{sec:filtering}

Despite the flexibility of 3DGS representations, direct optimization over the entire parameter space often produces artifacts such as isolated floating Gaussians and spatial jitter. These artifacts weaken physical plausibility, especially under multi-view rendering. We therefore propose a Physical Filtering Module to refine the optimized 3D object and enforce geometric consistency.

\vspace{1mm}
\noindent\textbf{Topological Pruning (TP).}
We first remove outlier Gaussians based on projected spatial density. For each Gaussian, we compute its neighborhood density in screen space across sampled viewpoints. If a Gaussian is consistently located in sparse regions or lacks sufficient overlap with other primitives, it is discarded. This step eliminates fragmented components and prevents unnatural floaters from appearing in rendered views.

\vspace{1mm}
\noindent\textbf{Structural Denoising (SD).}
We further smooth the object by adjusting the scale of neighboring Gaussians to reduce abrupt local variations. We apply a Gaussian-weighted smoothing filter over scale dimension, guided by projected image-space distance and opacity. This encourages gradual transitions in apparent size and reduces aliasing artifacts caused by irregular primitive distribution.

% Figure~\ref{fig:filtering} shows the effect of each filtering step. 

Formally, the filtering process involves (1) pruning low-density Gaussians
based on adaptive neighborhood analysis, and (2) applying a camera-aware
structural smoothing in scales and opacities.

Let $g_j$ denote the $j$-th Gaussian with center $x_j$.
We first estimate its local density $\rho(g_j)$ from its $k$-nearest neighbors 
in 3D space. An adaptive pruning threshold $\tau_d$ is obtained as the
$p$-th percentile of the density distribution:
\begin{equation}
    \tau_d = \mathrm{Quantile}\big(\{\rho(g_j)\}_{j=1}^{N},\, p\big).
\end{equation}
Gaussians satisfying $\rho(g_j) < \tau_d$ are treated as geometric outliers
and removed. This percentile-based rule automatically adapts to different
object densities without manual tuning of a fixed threshold.

For the remaining Gaussians, Structural Denoising applies a camera-aware
anisotropic filter over the scale parameters $s_j$ and opacities $\alpha_j$.
The filtering intensity $\sigma_j$ is modulated by the minimum projected
distance to all camera planes:
\begin{equation}
    \sigma_j \propto 
    \min_{i} \frac{\lVert R_i x_j + T_i \rVert_2^2}{f_i},
\end{equation}
where $(R_i, T_i, f_i)$ denote the rotation, translation, and focal length
of the $i$-th camera. This design enforces stronger smoothing in distant
regions while preserving fine details for closer views.

The filtered parameters are then obtained as
\begin{equation}
    s'_j = \sqrt{s_j^2 + \sigma_j^2 I}, 
    \qquad 
    \alpha'_j = \alpha_j \cdot 
    \frac{\det(s_j)}{\det(s'_j)}.
\end{equation}
Together, density-aware pruning and camera-aware smoothing suppress
floating artifacts and enforce multi-view geometric coherence, yielding
a refined 3DGS object $G_{\mathrm{base}}$ that serves as a physically 
plausible starting point for adversarial optimization.

\subsection{Physical Augmentation Module}
\label{sec:physical augmentation module}

To improve the deployment-time transferability of adversarial 3D objects, we introduce a differentiable \textit{Physical Augmentation Module} that simulates deployment-time environmental variations. This module injects physically realistic perturbations into rendered views during optimization and is composed of four sub-transforms:

\textit{Imaging Degradation.}
We simulate sensor noise and lens blur through depth-aware Gaussian noise $\mathcal{T}_{noise}$, where the noise level increases with object-camera distance. This models degradation from low-quality hardware or atmospheric scattering.

\textit{Photometric Variation.}
We apply color distortions via channel-wise affine transforms $\mathcal{T}_{photo}$, introducing random brightness and contrast changes to emulate lighting inconsistencies, exposure drift, and white balance variation.

\textit{Shadow Projection.}
To account for natural illumination effects, we simulate soft shadow overlays using depth-based sigmoid masks $\mathcal{T}_{shadow}$. These synthetic shadows mimic occlusion by other objects or self-casting.

\textit{Adaptive Occlusion.}
We introduce random rectangular occluders with variable transparency $\mathcal{T}_{occl}$ to simulate scene clutter and partial obstructions, enhancing robustness to unexpected foreground elements.

Each transformation is applied in sequence to the rendered image: $\mathcal{T} = \mathcal{T}_{occl} \circ \mathcal{T}_{shadow} \circ \mathcal{T}_{photo} \circ \mathcal{T}_{noise}$. All operations are differentiable and applied online at every training epoch, allowing gradients to flow through the augmentation pipeline. This enables adversarial optimization to produce samples that remain effective under realistic sensing and environmental distortions. Additional formulation details are provided in Appendix~\ref{sup: physical augmentation module}.

We further treat input preprocessing defenses as transformations applied between rendering and detector inference. In this setting, the adversarial 3DGS object remains fixed, and the detector receives a preprocessed version of the rendered image, such as a compressed, resized, blurred, or photometrically normalized input. This design tests whether the attack survives common input-space distortions without re-optimizing against the defense. It also clarifies the role of the Physical Augmentation Module: augmentation is not a post-hoc defense bypass, but a training-time exposure to nuisance factors that are likely to appear before detection.

The defense evaluation in Sec.~\ref{sec:defense_evaluation} is therefore a non-adaptive robustness check rather than a claim of complete security against all robust models. Preprocessing can remove fragile high-frequency image artifacts, but an object-level 3DGS perturbation may persist when the adversarial effect is carried by stable surface appearance, projected shape cues, and multi-view consistency. This boundary keeps the evaluation tied to the mechanism tested in the paper while leaving adversarially trained detectors and adaptive defenses as stronger future threat models.

\subsection{Dimension Selection for Physical Realization}
\label{sec:selective dimension optimization}
Vehicle deployment imposes a non-deformability constraint. To explicitly accommodate this, we expose a \emph{dimension-selection} switch in 3DGAA that partitions the 3DGS parameters into \(\theta_{\text{app}}=\{\text{rgb},\text{opacity}\}\) (appearance) and \(\theta_{\text{geo}}=\{\text{pos},\text{scale},\text{rot}\}\) (geometry), and instantiate three modes:

\noindent\textbf{Appearance-only $\text{3DGAA}^{a}$.} We \emph{freeze} \(\theta_{\text{geo}}\) and optimize only \(\theta_{\text{app}}\). In this setting, the resulting adversarial object can be fabricated by printable wraps/films without any geometric alteration (see Physical Realization). Technically, this mode achieves the best perceptual fidelity among attack methods (Table~\ref{tab:benchmark}: lowest LPIPS and highest PSNR).

\noindent\textbf{Geometry-only $\text{3DGAA}^{g}$.} We fix \(\theta_{\text{app}}\) and optimize \(\theta_{\text{geo}}\) under tight displacement bounds implicitly encoded by \(L_{\text{shape}}\). This ablates the contribution of geometric cues and keeps competitive realism (Table~\ref{tab:benchmark}) while exposing how pose/scale perturbations contribute to detector failures.

\noindent\textbf{Full (\textbf{3DGAA}).} We jointly optimize \(\theta_{\text{app}}\cup\theta_{\text{geo}}\) with \(L_{\text{shape}}\). This configuration provides the strongest attack effect (highest LCR, lowest mAP in Table~\ref{tab:benchmark}) at a small realism cost relative to $\text{3DGAA}^{a}$). 

The switchable design aligns optimization with physical constraints: $\text{3DGAA}^{a}$ directly supports geometry-free, printable deployment on vehicles; $\text{3DGAA}^{g}$ isolates geometric sensitivity; and \textbf{3DGAA} serves as an upper-bound setting for attack efficacy in our controlled evaluation. In practice, $\text{3DGAA}^{g}$ converges faster and with lower memory due to frozen appearance, while \textbf{3DGAA} achieves the best robustness. Moreover, $\text{3DGAA}^{a}$ serves as the methodological precursor to our fabrication pipeline (see Physical Realization) and the comprehensive comparisons in Exp.~\ref{sec:comprehensive comparison} and \ref{sec:physical realization}.

\begin{algorithm}[t]
\caption{Adversarial 3DGS Optimization}
\label{alg:adv-opt}
\begin{algorithmic}[1]
\small
\Require  
  \begin{tabular}[t]{@{}ll@{}}
    $T$: & Total iterations \\ 
    $\theta_0$: & Initial 3DGS parameters \\
    $R(\cdot)$: & 3DGS differentiable renderer \\
    $\mathcal{T}(\cdot)$: & Physical augmentation function \\
    $\mathcal{L}_{\text{adv}}$: & Adversarial loss \\
    $\mathcal{L}_{\text{shape}}$: & Shape consistency loss \\
    $\mathcal{K}$: & Selected dimensions for update
  \end{tabular}
\Ensure Optimized 3DGS parameters $\theta$

\State $\theta \gets \theta_0$ \Comment{Initialization}
\For{$t = 1$ \textbf{to} $T$}
    \State $I \gets R(\theta)$ \Comment{Render current 3DGS}
    \State $\tilde{I} \gets \mathcal{T}(I)$ \Comment{Apply physical augmentation}
    \State Compute weights $\lambda_{\text{adv}}, \lambda_{\text{shape}}$ based on current loss values
    \State $\delta \gets \nabla_{\theta} \left[ \lambda_{\text{adv}} \cdot \mathcal{L}_{\text{adv}}(\tilde{I}) + \lambda_{\text{shape}} \cdot \mathcal{L}_{\text{shape}}(\theta) \right]$
    \State $\theta \gets \theta - \text{Mask}_{\mathcal{K}}(\delta)$ \Comment{Update selected dimensions only}
\EndFor
\State \textbf{return} $\theta$
\end{algorithmic}
\end{algorithm}

%% file: sec/04_experimentsc.tex
\section{Experiments}
\label{sec:exp}

\subsection{Experimental Setup}

\noindent\textbf{Baseline Methods.}
We compare 3DGAA against representative 2D and 3D physical attack baselines, including CAMOU~\cite{Zhang_Foroosh_David_Gong_2018}, UPC~\cite{Huang_Gao_Zhou_Xie_Yuille_Zou_Liu_2020}, DAS~\cite{Wang_Liu_Yin_Liu_Tang_Liu_2021}, FCA~\cite{Wang_Jiang_Sun_Zhou_Gong_Zhang_Yao_Chen_2022}, DTA~\cite{Suryanto_Kim_Kang_Larasati_Yun_Le_Yang_Oh_Kim}, ACTIVE~\cite{suryanto2023activehighlytransferable3d}, TT3D~\cite{huang2024towards}, RAUCA~\cite{RAUCA} and PACG~\cite{PACG}.
For a fair comparison, all methods are adapted to the same synthetic and controlled physical setups with standard calibration and consistent rendering settings.

\vspace{1mm}
\noindent\textbf{Object \& Invariance.}
\revise{All physical variants use the \emph{same rigid miniature shell}; geometry is unchanged and only textures differ (Sec.~\ref{sec:physical realization}).
Textures used for fabrication are produced by the \textbf{appearance-only} pipeline \(\text{3DGAA}^{a}\) (Sec.~\ref{sec:selective dimension optimization}).}{All physical variants in Table~\ref{tab:physical comparison} use the same rigid miniature shell and capture protocol; geometry remains fixed, and only the applied texture differs. The 3DGAA wrap is generated by the \textbf{appearance-only} pipeline \(\text{3DGAA}^{a}\) (Secs.~\ref{sec:selective dimension optimization} and~\ref{sec:physical realization}).}
\revnote{AE/R1: Clarified the invariant shell and matched capture protocol used for the controlled physical baseline comparison.}

\vspace{1mm}
\noindent\textbf{Synthetic Scene Selection.}
We adopt CARLA~\cite{dosovitskiy2017carlaopenurbandriving} sampling with fixed seeds to collect the images, covering 20 vehicle models with varying geometries, 5 weather conditions, multi-scale observations and full spherical viewpoints (24 azimuth/elevation angles), 12000 images in total.

\begin{table*}[t!]
    \centering
    % \caption{\textbf{Benchmark Comparison} of 3DGAA against existing physical adversarial attacks. Across metrics, our approach consistently delivers superior physical realism and adversarial robustness.}
    \caption{\textbf{Benchmark Comparison} of 3DGAA against existing physical adversarial attacks. Besides the full model (3DGAA), we report two dimension-selection variants: $\text{3DGAA}^{a}$ (appearance-only optimization) and $\text{3DGAA}^{g}$ (geometry-only optimization). Across metrics, our approach delivers strong physical realism and adversarial robustness among the evaluated methods.}
    \vspace{-2mm}
    \setlength{\tabcolsep}{2.7pt}
    \footnotesize                 
    \begin{tabular}{lccccccccccccc}
        \toprule
        Method  & Vanilla & CAM.\cite{Zhang_Foroosh_David_Gong_2018} & UPC\cite{Huang_Gao_Zhou_Xie_Yuille_Zou_Liu_2020} & DAS\cite{Wang_Liu_Yin_Liu_Tang_Liu_2021} & FCA\cite{Wang_Jiang_Sun_Zhou_Gong_Zhang_Yao_Chen_2022} & DTA\cite{Suryanto_Kim_Kang_Larasati_Yun_Le_Yang_Oh_Kim} & ACT.\cite{suryanto2023activehighlytransferable3d} & TT3D\cite{huang2024towards} & RAU.\cite{RAUCA} & PACG\cite{PACG} & $\textbf{3DGAA}^a$ & $\textbf{3DGAA}^g$ & \textbf{3DGAA}\\ 
        \toprule
        LPIPS $\downarrow$&  0.0000 & 0.6210 & 0.6161 & 0.5979 & 0.5629 & 0.6142 & 0.6105 & 0.5433 & 0.6056 & 0.6152 & \best{0.5218} & \thirdbest{0.5423} & \secondbest{0.5373} \\
        SSIM  $\uparrow$  &  1.0000 & \secondbest{0.0212} & \best{0.0224} & 0.0132 & 0.0123 & 0.0198 & 0.0193 & 0.0122 & 0.0178 & \thirdbest{0.0204} & 0.0123 & 0.0193 & 0.0128 \\
        PSNR  $\uparrow$  &  $\infty$&0.1829 & 0.1749 & 0.4590 & 0.2987 & 0.1804 & 0.1814 & 0.3876 & 0.2572 & 0.1956 & \best{0.5480} & \thirdbest{0.4741} & \secondbest{0.4951}\\
        LCR   $\uparrow$  &  0.0000  & 0.3128  & 0.2095  & 0.5366  & 1.2810 & 0.8784 & 1.3778 & 1.1202 & 1.4705 & 1.0299 & \thirdbest{2.6766} & \secondbest{3.4410} & \best{3.5628}\\
        mAP   $\downarrow$&  87.21\%& 70.21\% & 75.42\% & 60.12\% & 35.89\%& 47.44\%& 33.56\%& 40.12\% & 31.47\% & 42.71\% & \thirdbest{13.64\%} & \secondbest{8.03\%} & \best{7.38\%}\\
        ASR  $\uparrow$ &  1.33\%   & 21.44\%   & 16.15\%    & 38.38\%   & 51.18\% & 47.77\% & 61.35\% & 56.30\% & 63.54\% & 52.90\% & 77.32\% & 85.68\% & 91.39\% \\
        \bottomrule
    \end{tabular}

    \label{tab:benchmark}
\end{table*}

% \begin{table}
%     \centering
%     \caption{\textbf{Ablation Study} on different optimization dimensions of 3DGAA. We compare the full model (3DGAA) with two dimension-selection variants: $\text{3DGAA}^{a}$ (appearance-only optimization) and $\text{3DGAA}^{g}$ (geometry-only optimization).}\vspace{-2mm}
%     \footnotesize                 
%     \begin{tabular}{lccc}
%         \toprule
%         Method & $\textbf{3DGAA}^a$ & $\textbf{3DGAA}^g$ & \textbf{3DGAA}\\
%         \midrule
%         LPIPS $\downarrow$ & \best{0.5218} & \thirdbest{0.5423} & \secondbest{0.5373} \\
%         SSIM $\uparrow$ & 0.0123 & 0.0193 & \best{0.0128} \\
%         PSNR $\uparrow$ & \best{0.5480} & \thirdbest{0.4741} & \secondbest{0.4951} \\
%         LCR $\uparrow$ & \thirdbest{2.6766} & \secondbest{3.4410} & \best{3.5628} \\
%         mAP $\downarrow$ & \thirdbest{13.64\%} & \secondbest{8.03\%} & \best{7.38\%} \\
%         ASR $\uparrow$ & 77.32\% & 85.68\% & 91.39\%\\
%         \bottomrule
%     \end{tabular}
%     \label{tab:benchmark_ablation}
% \end{table}
\begin{table}[t]
    \centering
    \caption{Attack performance of 3DGAA on CNN-based and transformer-based detectors in simulation.}
    \label{tab:transformer_detector}
    \setlength{\tabcolsep}{4pt}
    \small
    \begin{tabular}{lccccc}
        \toprule
        Detector & Type & Clean mAP & Attack mAP $\downarrow$   &LCR $\uparrow$ & ASR $\uparrow$\\
        \midrule
        YOLOv3 & CNN & 87.21\% & 7.38\%& 3.5628  & 91.39\% \\
        SSD & CNN & 60.25\% & 7.92\% & 2.9274 & 71.74\% \\
        F-RCNN & CNN & 71.06\% & 8.59\% & 3.0483 & 88.64\% \\
        M-RCNN & CNN & 64.75\% & 5.23\% & 3.6300 & 81.96\% \\
        D-DETR & Tf. & 36.57\% & 6.51\% & 2.4899 & 69.20\% \\
        DINO & Tf. & 51.96\% & 30.24\% & 0.7810 & 80.21\% \\
        \bottomrule
    \end{tabular}
    % \vspace{1mm}
    % \parbox{\linewidth}{\raggedright \scriptsize \vspace{0.5mm}Tf.: transformer-based detectors; D-DETR: Deformable DETR.}
    \parbox{\linewidth}{\vspace{0.5mm}\small\raggedright Tf.: transformer-based detectors; D-DETR: Deformable DETR.}

\end{table}

% \vspace{1mm}
\noindent\textbf{Target Models.}
% We evaluate detection attacks on widely used CNN-based detectors, including Faster R-CNN~\cite{frcnn}, Mask R-CNN~\cite{mrcnn}, SSD~\cite{ssd}, YOLOv3~\cite{yolov3} and YOLOv5, and further include two transformer-based detectors, Deformable DETR (D-DETR)~\cite{ddetr} and DINO~\cite{dino}.
We evaluate detection attacks on widely used CNN-based detectors, including Faster R-CNN~\cite{frcnn}, Mask R-CNN~\cite{mrcnn}, SSD~\cite{ssd}, YOLOv3~\cite{yolov3} and YOLOv5, and further include two transformer-based detectors, Deformable DETR (D-DETR)~\cite{ddetr} and DINO~\cite{dino}.

For transferable task segmentation, we use DeepLabv3~\cite{chen2017rethinkingatrousconvolutionsemantic} and FCN~\cite{fcn} with ResNet-50 and ResNet-101~\cite{he2015deepresiduallearningimage} backbones.

\vspace{1mm}
\noindent\textbf{Evaluation Metrics.}
We report attack effectiveness using the \textbf{Log Confidence Reduction (LCR)}, which measures the log-ratio reduction from the initial detector confidence to the final detector confidence after attack:
\begin{equation}
\label{equ:lcr}
\text{LCR}=\log\!\left(\frac{\text{Initial Confidence}}{\text{Final Confidence}}\right).
\end{equation}
Unlike absolute confidence drop, LCR normalizes adversarial effectiveness across different baselines, ensuring a fairer comparison.
We additionally report Attack Success Rate (\textbf{ASR}) for detection attacks. ASR is the percentage of test images in which the target object is no longer correctly detected after attack under the same confidence and IoU matching protocol used for mAP evaluation.
We also report mean Average Precision with 50\% threshold (\textbf{mAP@0.5}) to reflect detection performance decline.
For semantic segmentation, we replace confidence-based LCR with mean Intersection-over-Union (\textbf{mIoU}), which is the standard task-level metric for segmentation quality. We report clean mIoU, adversarial mIoU, and mIoU Drop, where mIoU Drop is the absolute decrease from clean mIoU to adversarial mIoU under the same view protocol.

To quantify physical realism, we adopt image similarity metrics including \textbf{LPIPS}~\cite{zhang2018perceptual}, \textbf{SSIM}~\cite{SSIM}, and \textbf{PSNR}, computed between rendered adversarial objects and their clean counterparts.
All metrics are computed under the same view and lighting protocol as in Sec.~\ref{sec:comprehensive comparison} to ensure fairness in a shared CAV-like setting.

\vspace{1mm}
\noindent\textbf{Implementation Details.}
All experiments run on a single NVIDIA RTX~4090 with fixed driver/CUDA/PyTorch versions and fixed seeds.
We use early stopping under a shared rule; \emph{minutes-level} wall-clock runtime profiling and peak memory under this protocol are reported in Table~\ref{tab:efficiency}.
The two loss terms in our objective--adversarial loss \(\mathcal{L}_{\text{adv}}\) and shape-consistency loss \(\mathcal{L}_{\text{shape}}\)--are dynamically weighted according to the scheme in Sec.~\ref{sec:optimization}, eliminating manual tuning of fixed loss weights.

\vspace{1mm}
\noindent\textbf{Generation Pipeline and Camera Setup.}
\emph{Digital rendering:} we use a differentiable 3D Gaussian renderer following~\cite{tang2024lgmlargemultiviewgaussian}.
All adversarial objects are rendered under multi-view settings with 12 evenly spaced azimuth angles and 3 distances (3\,m, 5\,m, 10\,m), aligned with the protocol in Sec.~\ref{sec:comprehensive comparison}.
\emph{Physical fabrication:} textures are UV-unwrapped and printed on \emph{matte wraps} to reduce glare, then \emph{applied to the same rigid shell} (no body modification); materials, printing ppi, and alignment cues are summarized in App.~\ref{app:implementation details}.
Physical images are captured with a Realme GT~Neo5~SE; camera/exposure specifics and layout sketches are provided in App.~\ref{app:implementation details}.

\subsection{Comprehensive Comparison with Benchmark Physical Attacks}
\label{sec:comprehensive comparison}
We compare 3DGAA with a range of representative physical adversarial attack methods, including both 2D patch-based and 3D texture-based approaches. Our evaluation considers two key dimensions: \textit{physical realism}, how visually consistent the adversarial object is with its original counterpart, and \textit{attack effectiveness}, how successfully it deceives object detection models in camera-based autonomous driving perception pipelines.

\vspace{1mm}
\noindent\textbf{Physical Realism.}
As shown in Table~\ref{tab:benchmark}, 3DGAA\textsuperscript{a} attains the lowest LPIPS (0.5218), with 3DGAA (0.5373) and 3DGAA\textsuperscript{g} (0.5423) following as second and third best, respectively. While our SSIM is marginally lower than UPC, this largely reflects the inherent approximation of 3DGS geometry rather than adversarial manipulation. Importantly, 3DGAA\textsuperscript{a} achieves the highest PSNR (0.5480) among attack methods, with 3DGAA second (0.4951) and 3DGAA\textsuperscript{g} third (0.4741), indicating stronger pixel-level fidelity among the evaluated attack methods. These results highlight the benefit of our shape-consistency loss and physical filtering in preserving realism, and show that appearance-only optimization can further tighten perceptual fidelity.

\vspace{1mm}
\noindent\textbf{Attack Strength.}
3DGAA significantly outperforms all baselines in adversarial robustness (Table~\ref{tab:benchmark}). It achieves an LCR of 3.5628, exceeding the strongest baseline RAUCA (1.4705), and reduces mAP from 87.21\% to 7.38\%. The ASR results show the same trend: full 3DGAA reaches 91.39\% ASR, compared with 63.54\% for the strongest non-3DGAA baseline. Both variants also surpass all baselines: 3DGAA\textsuperscript{g} reaches 3.4410 LCR / 8.03\% mAP and 3DGAA\textsuperscript{a} 2.6766 LCR / 13.64\% mAP. This confirms that optimizing both geometry and appearance yields the strongest effect, while our selective dimension optimization (Sec.~\ref{sec:selective dimension optimization}) still delivers SOTA degradation.

\vspace{1mm}
\noindent\textbf{Perturbation-Area Fairness.}
Because local patch attacks and object-level attacks operate under different perturbation scopes, Table~\ref{tab:perturbation_area} adds a fairness-aware perturbation-area analysis in rendered views. For each clean/adversarial rendering pair, we compute PAR@1/255 as the percentage of pixels whose mean RGB absolute difference exceeds \(1/255\). We also report the mean absolute difference (MAD) and the 95th-percentile absolute difference (P95) on the /255 RGB scale.

\begin{table}[t]
    \centering
    \caption{Fairness-aware perturbation-area analysis in rendered views. PAR@1/255 reports the percentage of pixels whose mean RGB absolute difference exceeds \(1/255\). MAD and P95 are reported on the /255 RGB scale.}
    \label{tab:perturbation_area}
    \setlength{\tabcolsep}{1.6pt}
    \footnotesize
    \begin{tabularx}{\columnwidth}{l>{\centering\arraybackslash}p{0.72cm}ccccc}
        \toprule
        Method & Scope & PAR@1/255 $\downarrow$ & MAD $\downarrow$ & P95 $\downarrow$ & LCR $\uparrow$ & ASR $\uparrow$ \\
        \midrule
        UPC & L-2D & 5.03$\pm$1.02\% & 0.47$\pm$0.22 & 15.69$\pm$5.05 & 0.21 & 16.15\% \\
        DAS & L-2D & 6.35$\pm$2.62\% & 0.41$\pm$0.16 & 17.39$\pm$6.22 & 0.54 & 38.38\% \\
        FCA & 3D-T & 26.45$\pm$8.90\% & 2.70$\pm$0.91 & 37.22$\pm$8.61 & 1.28 & 51.18\% \\
        RAUCA & 3D-T & 32.57$\pm$11.36\% & 2.86$\pm$1.18 & 40.26$\pm$10.89 & 1.47 & 63.54\% \\
        \textbf{3DGAA} & 3DGS & 8.99$\pm$1.45\% & 0.53$\pm$0.14 & 3.37$\pm$1.14 & 3.56 & 91.39\% \\
        \bottomrule
    \end{tabularx}
    \parbox{\linewidth}{\vspace{0.5mm}\small\raggedright  L-2D: local 2D perturbation; 3D-T: 3D texture attack.}
\end{table}

The area analysis clarifies the comparison boundary rather than replacing the main attack benchmark. Local 2D attacks naturally concentrate their changes in a small region, which leads to low perturbation area but strong local pixel changes. Texture-based 3D attacks affect broader surface regions and therefore introduce more visible appearance changes. In contrast, 3DGAA is optimized as an object-level 3DGS representation, but its rendered changes remain low in magnitude. This indicates that its attack strength comes from coordinated geometry-aware and appearance-aware changes in the 3D representation, not from simply enlarging a visible patch or repainting a large surface area.

\vspace{1mm}
\noindent\textbf{Transformer-Based Detectors.}
To examine whether 3DGAA remains effective beyond CNN-based detectors, we additionally evaluate D-DETR and DINO as target detectors under the same simulation protocol. Table~\ref{tab:transformer_detector} reports clean mAP, attacked mAP, LCR, and ASR for both CNN-based and transformer-based detectors. 3DGAA reduces D-DETR mAP from 36.57\% to 6.51\% and achieves 69.20\% ASR. On DINO, the attacked mAP decreases from 51.96\% to 30.24\%, with 80.21\% ASR. These results show that the attack is not limited to CNN-style detection pipelines, although the magnitude of degradation varies across detector architectures. Because the transformer-based detectors start from lower clean-target mAP in this protocol, their post-attack numbers should be read together with ASR and LCR rather than as a direct claim of larger absolute degradation.

\begin{figure}[t]
    \centering
    \includegraphics[width=0.65\linewidth]{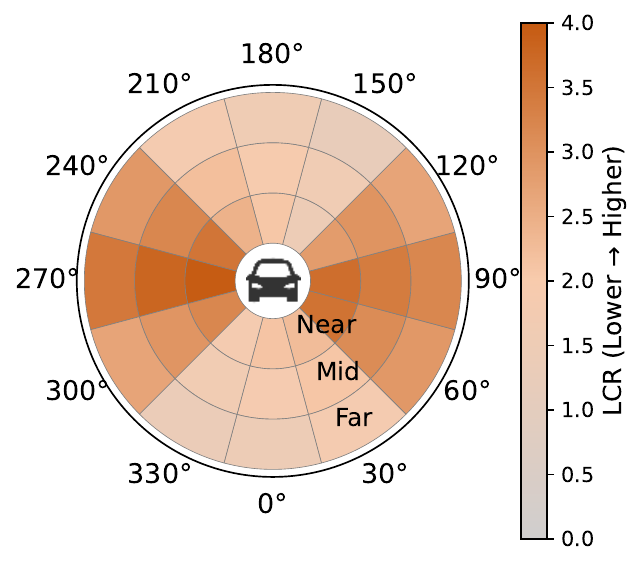}\vspace{-3mm}
    \caption{Polar plot of LCR across 12 viewpoints and 3 distances in simulation. Attacks are most effective at side views and short range.}\vspace{-3mm}
    \label{fig:polar}
\end{figure}

\vspace{1mm}
\noindent\textbf{Viewpoint Robustness.}
We measure LCR across 12 azimuth angles and 3 distances. Figure~\ref{fig:polar} shows higher success rates at side views and close range, with performance slightly declining under front views or long distances, validating the multi-view training strategy and 3DGAA's robustness to viewpoint variations. Extended multi-view visualizations in the supplementary material (Figs.~\ref{fig:sup_orig} and~\ref{fig:sup_adv}) further show that geometric structure is preserved while detections are consistently suppressed across viewpoints.

\vspace{1mm}
\noindent\textbf{Human-Perception Study.}
To complement objective metrics, we conduct a human-perception study with 50 volunteers. Each participant is presented with a randomized mix of vanilla and adversarial textures rendered on vehicles, and asked to rate perceived realism on a 1--10 scale (10: indistinguishable from a natural paint scheme; 1: clearly artificial or adversarial).
Table~\ref{tab:human_perception} reports the mean realism scores.
Vanilla textures score 9.6, while naive random patterns obtain 1.1.
Among adversarial baselines, UPC and ACTIVE reach 6.4 and 6.0, respectively, whereas CAMOU, DAS and FCA remain below 4.0 due to large, obviously artificial distortions.
Our 3DGAA achieves the highest realism score (7.9), and the appearance-only variant $\text{3DGAA}^{a}$ also scores highly (7.1), both much closer to vanilla than to other attacks.
This indicates that 3DGAA produces perturbations that remain largely imperceptible to humans, a desirable property for realistic security and safety assessment of camera-based perception.

\begin{table}[t]
    \centering
    \caption{Realism scores of different adversarial and vanilla textures from participants.}\vspace{-2mm}
    \label{tab:human_perception}
    \begin{tabular}{lclc}
        \toprule
        Methods & Scores & Methods & Scores\\
        \midrule
        Vanilla & 9.6 & Random & 1.1 \\
        CAMOU & 2.9 & UPC& 6.4  \\
        DAS & 3.9 & FCA & 2.3 \\
        DTA & 3.2 & ACTIVE & 6.0 \\
        RAUCA & 3.7 & PACG & 3.6 \\
        $\text{3DGAA}^a$ & 7.1 & 3DGAA & 7.9\\
        \bottomrule
    \end{tabular}
    
    \vspace{0.5mm}    
    \small\raggedright $\text{3DGAA}^a$: appearance-only mode.
\end{table}

Together, these results confirm that 3DGAA achieves strong attack success while preserving physical realism, and remains effective under diverse viewpoints. Moreover, the strong results of \textbf{3DGAA}\textsuperscript{a} verify a purely appearance-based and geometry-free deployment path, which is critical for physically deployable adversarial objects in autonomous scenarios.

% \vspace{-10pt}

\subsection{Effectiveness of the Physical Filtering Module}
\begin{figure}[t]
    \centering
    \includegraphics[width=0.95\linewidth, trim=0 0 0 0, clip]{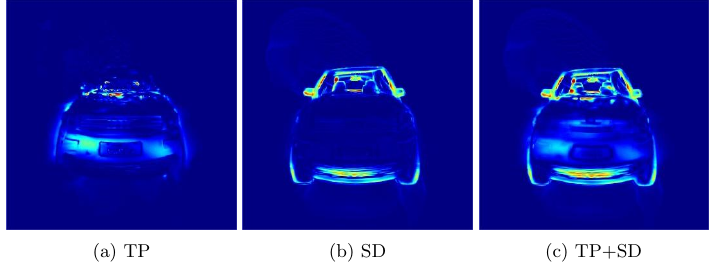}\vspace{-2mm}
    \caption{\textbf{Geometric refinement analysis through physical filtering module}: (a) Topology pruning (b) Structural denoising (c) Topology pruning + Structural denoising. Color scale indicates local deformation energy.}\vspace{-3mm}
    \label{fig:filter_heatmap}
\end{figure}

To evaluate the impact of the physical filtering module introduced in Section~\ref{sec:filtering}, we compare the results of applying topology pruning (TP), structural denoising (SD), and their combination. As shown in Figure~\ref{fig:filter_heatmap}, the visual artifacts, such as floating Gaussians and irregular surface patches, are significantly reduced after filtering. The TP step removes geometric outliers, while SD smooths local scale inconsistencies. The combined TP+SD strategy yields the cleanest geometry and most plausible object structure.

Quantitatively, Table~\ref{tab:filtering_table} shows that TP+SD improves perceptual realism (LPIPS: 0.5373), achieves the highest SSIM (0.0128) and removes 83.1\% of artifacts, although with a slight decrease in PSNR due to suppression of high-frequency noise. This trade-off is acceptable, as it prioritizes geometric plausibility over strict pixel-level fidelity, aligning with our goal of producing physically realistic adversarial objects.

\begin{table}[t]
    \centering
    \caption{Filtering strategy comparison across different metrics. TP+SD achieves the best performance on physical realism and artifacts removal (AR).}\vspace{-2mm}
    \label{tab:filtering_table}
    \begin{tabular}{lcccc}
        \toprule
        Method & LPIPS $\downarrow$ & SSIM $\uparrow$ & PSNR $\uparrow$ & AR (\%) $\uparrow$ \\
        \midrule
        Vanilla     & 0.5443 & 0.0105 & \textbf{0.5282} & 0.0 \\
        TP          & 0.5435 & 0.0086 & 0.5251 & 19.8 \\
        SD          & 0.5386 & 0.0121 & 0.5015 & 72.2 \\
        TP + SD     & \textbf{0.5373} & \textbf{0.0128} & 0.4951 & \textbf{83.1} \\
        \bottomrule
    \end{tabular}\vspace{-2mm}
\end{table}

\begin{table}[t]
    \centering
    \caption{Ablation study of the shape loss component. $\mathcal{L}_{adv} + \mathcal{L}_{shape}$ achieves better physical realism.}\vspace{-2mm}
    \label{tab:shape_loss}
    \begin{tabular}{lccc}
        \toprule
        Method & LPIPS $\downarrow$ & SSIM $\uparrow$ & PSNR $\uparrow$ \\
        \midrule
        $\mathcal{L}_{adv}$ only                 & 0.5516 & 0.0141 & \textbf{0.5884} \\
        $\mathcal{L}_{adv} + \mathcal{L}_{shape}$ & \textbf{0.5503} & \textbf{0.0146} & 0.5881 \\
        \bottomrule
    \end{tabular}\vspace{-2mm}
\end{table}

\begin{table}[t]
    \centering
    \caption{Ablation study of the Physical Filtering and Physical Augmentation modules.}
    \label{tab:module_ablation}
    \setlength{\tabcolsep}{4pt}
    \small
    \begin{tabular}{lcccc}
        \toprule
        Method & LCR $\uparrow$ & mAP $\downarrow$ & LPIPS $\downarrow$ & PSNR $\uparrow$ \\
        \midrule
        Full 3DGAA & \best{3.5628} & \best{7.38\%} & \best{0.5373}  & \best{0.4951} \\
        w/o Phy. Fil. & \secondbest{2.7854} & \secondbest{12.65\%} & 0.5823 & 0.3736 \\
        w/o Phy. Aug. & 1.2926 & 35.60\% & \secondbest{0.5548} & \secondbest{0.4629} \\
        w/o Both & 1.0496 & 42.13\% & 0.5962 & 0.3403 \\
        \bottomrule
    \end{tabular}
\end{table}

\subsection{Module Ablation}
We further ablate the two physical modules jointly to examine whether their benefits come only from metric-level improvements or from distinct roles in deployable attack generation. As shown in Table~\ref{tab:module_ablation}, removing either module degraded the full system, and removing both modules produced the weakest result. Full 3DGAA achieved the highest LCR (3.5628), the lowest mAP (7.38\%), and the best perceptual realism among the four variants. This confirms that Physical Filtering and Physical Augmentation are complementary rather than redundant.

The effect of removing Physical Filtering is especially informative. The variant without Physical Filtering still retained part of the adversarial capability, with LCR decreasing from 3.5628 to 2.7854 and mAP increasing from 7.38\% to 12.65\%. However, its LPIPS and PSNR degraded substantially, indicating that part of the attack effect was carried by poorly supported Gaussians, floating artifacts, and high-frequency surface noise. Such perturbations can reduce detector confidence in rendered images, but they are difficult to preserve through printing, wrapping, or real-object deployment. Physical Filtering therefore acts as a realism-preserving constraint that redirects the attack from non-fabricable artifacts toward physically coherent object surfaces.

In contrast, removing Physical Augmentation preserved part of the rendering realism but sharply weakened attack effectiveness. The w/o Physical Augmentation variant maintained relatively competitive LPIPS and PSNR, yet its LCR dropped to 1.2926 and its mAP rose to 35.60\%. This indicates that photorealistic rendered appearance alone is insufficient for robust physical attack performance. Without exposure, lighting, noise, and occlusion variation during optimization, the learned perturbation overfits to nominal rendering conditions and loses much of its adversarial effect under deployment-time nuisance factors. This trend is further supported by the physical realization results in Sec.~\ref{sec:physical realization}, where Fig.~\ref{fig:physical_realization} shows that the non-augmented object was sensitive to illumination changes, whereas the augmented 3DGAA object remained effective across different conditions.

\begin{figure}[t]
    \centering
    \begin{subfigure}[b]{0.48\linewidth}
        \includegraphics[width=\linewidth]{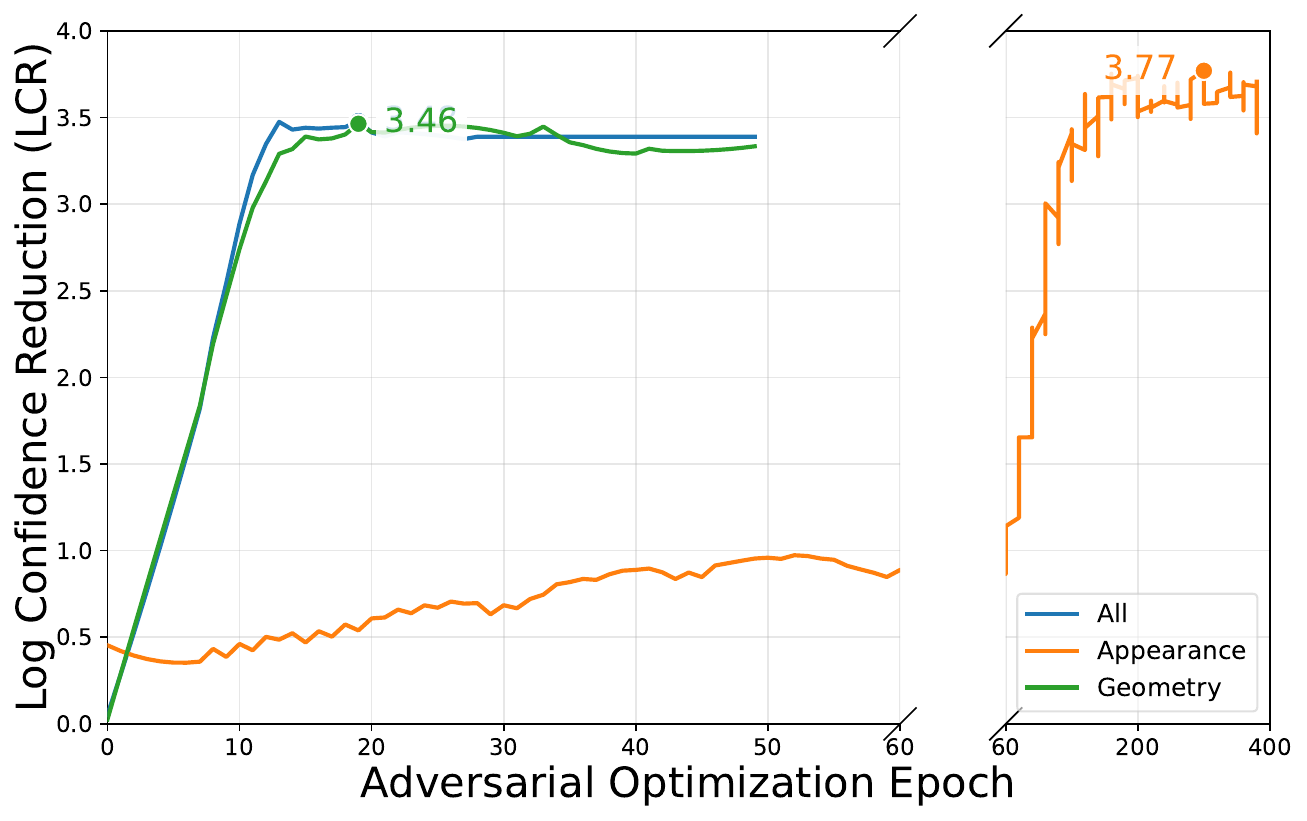}
        \caption{Grouped optimization.}
        \label{fig:part}
    \end{subfigure}
    \hfill
    \begin{subfigure}[b]{0.48\linewidth}
        \includegraphics[width=\linewidth]{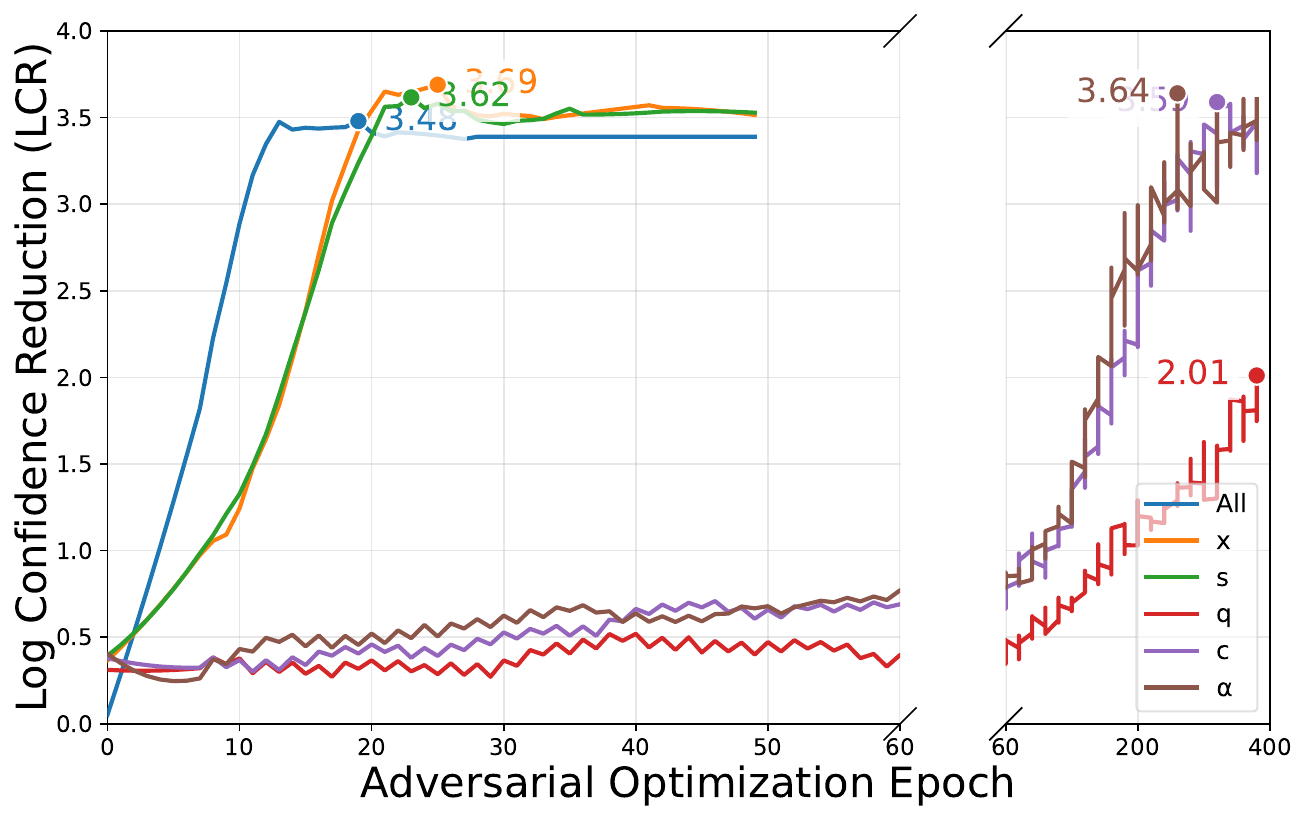}
        \caption{Single-Dim optimization.}
        \label{fig:all}
    \end{subfigure}\vspace{-2mm}

    \caption{LCR sensitivity analysis. (a) Optimizing geometry dimensions leads to faster convergence, while texture yields higher final LCR. (b) Position and scale are most sensitive; rotation has minimal effect. \protect\footnotemark}
    \label{fig:dim_analysis}\vspace{-5mm}

\end{figure}

\subsection{Effect of Shape Loss Design}

We conduct an ablation study to assess the contribution of the shape loss $\mathcal{L}_{shape}$ introduced in Section~\ref{sec:optimization}. This loss aims to preserve geometric consistency while optimizing for adversarial effectiveness.

As shown in Table~\ref{tab:shape_loss}, adding the shape loss slightly improves LPIPS (0.5516 to 0.5503) and SSIM (0.0141 to 0.0146), indicating better perceptual quality and structural preservation. The PSNR value remains similar, with only a minor drop, reflecting a shift from pixel-level perturbation to more physically realistic adjustments.
These results show that incorporating $\mathcal{L}_{shape}$ helps balance attack robustness and physical realism, aligning with our dual-objective design.

\footnotetext{The broken axis does not indicate discontinuous data, but compresses the horizontal scale to better visualize post-60-epoch results.}

\subsection{Selective Dimension Optimization}
\label{sec:exp_selective}
To analyze the sensitivity and effectiveness of individual 3DGS parameters during adversarial optimization, we perform a set of experiments that isolate geometric and appearance dimensions.

As shown in Figure~\ref{fig:part}, optimizing geometric parameters (position $x$, scale $s$) leads to rapid increases in Log Confidence Reduction (LCR), achieving strong attack effectiveness within 30 epochs. In contrast, optimizing texture parameters (opacity $\alpha$, color $c$) requires more iterations but ultimately reaches comparable or higher LCR. This suggests a time-performance trade-off: geometry enables fast convergence, while texture offers stronger long-term perturbation due to its direct effect on visual semantics.

A detailed dimension-wise analysis in Figure~\ref{fig:all} further reveals that position $x$ and scale $s$ are the most sensitive parameters, contributing significantly to early-stage attack progress. Opacity $\alpha$ and color $c$ exhibit moderate sensitivity, as they influence visibility and texture realism, respectively. Rotation $q$, despite affecting global geometry, shows minimal impact, due to its non-local and redundant nature in Gaussian-based representations. These findings support our design of selective dimension optimization (Section~\ref{sec:selective dimension optimization}), which allows practitioners to balance computational efficiency and physical feasibility in deployment-oriented evaluations.

\vspace{1mm}
\noindent\textbf{Computational Efficiency.}
We further profile the runtime and memory footprint of different 3DGAA variants and two non-3DGS baselines under a shared protocol (same data, views, and early-stopping rule).
Table~\ref{tab:efficiency} reports wall-clock times from process start to end, decomposed into 3DGS generation, adversarial optimization, and data I/O/miscellaneous overhead, together with peak GPU memory.
Full 3DGAA converges within $51.52$\,s end-to-end, while $\text{3DGAA}^{g}$ and $\text{3DGAA}^{a}$ finish in $48.92$\,s and $116.32$\,s, respectively, on a single RTX~4090.
In contrast, FCA and TT3D operate at tens of seconds per iteration and hours-level totals.
These profiles indicate that the proposed 3DGAA family can be used as a practical stress-testing tool for camera-based perception stacks in offline security and robustness evaluation.

\begin{table*}[t]
\centering
\caption{Efficiency profiling under a shared protocol (same data, views, and early-stopping rule). 
Times are wall-clock from process start to end; \emph{Total} includes 3DGS generation, adversarial training, and data I/O/misc.}
\label{tab:efficiency}
\setlength{\tabcolsep}{3.2pt}
\small\vspace{-2mm}
\begin{tabular}{lcccccccccccc}
\toprule
Mode & GPU & Peak Mem & 3DGS Gen & Adv. Iters & Time/Iter & Adv. Time & Data/Misc & \textbf{Total} & G & $\Delta$G \\
 & (Model) & (MB) & (s) & (epochs/iters) & (s) & (s) & (s) & (s) &  & (\%)  \\
\midrule
\textbf{3DGAA} (Full)   & RTX 4090 & 23256 & $9.24\pm0.26$ & $27$ & $0.37$ & $10.02\pm0.22$ & $32.26\pm0.01$ & $51.52$ & $24378$  & $98.22$\\
$\textbf{3DGAA}^{a}$    & RTX 4090 & 19862 & $9.24\pm0.26$ & $283$ & $0.26$ & $74.82\pm3.20$ & $32.26\pm0.01$ & $116.32$ & $24378$ & $80.25$\\
$\textbf{3DGAA}^{g}$    & RTX 4090 & 21196 & $9.24\pm0.26$ & $24$ & $0.31$ & $7.42\pm0.15$ & $32.26\pm0.01$ & $48.92$ & $24378$  & $75.05$\\
FCA~\cite{Wang_Jiang_Sun_Zhou_Gong_Zhang_Yao_Chen_2022} & RTX 4090 & 23256 & -- & 200 & 12.68 & $2536\pm96$ & $92.52\pm12.18$ & 2629 & -- & --\\
TT3D~\cite{huang2024towards} & RTX 4090 & 23256 & -- & 200 & 21.80 & $4360\pm74$ & $166.29\pm48.40$ & 4529 & -- & --\\
\bottomrule
\end{tabular}\vspace{-5mm}
\end{table*}

\subsection{Transferability Across Models and Tasks}

We evaluate the transferability of 3DGAA adversarial objects across different perception models, including detectors with diverse architectures and segmentation models.

\vspace{1mm}
\noindent\textbf{Cross-Detector Generalization.}
Different from Table~\ref{tab:transformer_detector}, which evaluates each detector as the target model, Table~\ref{tab:cross-detector} examines whether an adversarial object optimized on one detector transfers to other detectors.
Table~\ref{tab:baseline_transferability} further compares 3DGAA with strong physical attack baselines under a shared transfer protocol, where all adversarial objects are optimized on YOLOv3 and tested on unseen detectors.
Perturbations generated from two-stage detectors (Faster R-CNN and Mask R-CNN) demonstrate strong transferability, reaching LCR = 6.180 on YOLOv3 and maintaining performance on SSD (e.g., F-RCNN $\rightarrow$ SSD yields LCR = 2.193). The baseline comparison shows the same trend from a method-level perspective: 3DGAA achieves the highest average transfer LCR (1.0579), exceeding RAUCA (0.8093), FCA (0.7115), TT3D (0.6929), and PACG (0.4845).
The SSD column remains informative because it receives lower transfer from some source models in Table~\ref{tab:cross-detector}, yet 3DGAA still preserves the strongest method-level transfer in Table~\ref{tab:baseline_transferability}.

These results suggest that 3DGAA captures transferable vulnerabilities across architectures through multi-dimensional optimization, rather than overfitting to a specific detector.

\begin{table}[t]
    \centering
    \caption{LCR ($\uparrow$) of 3DGAA adversarial objects optimized on one source detector and evaluated on different target detectors. Higher values indicate stronger transferability.}
    \label{tab:cross-detector}
    \setlength{\tabcolsep}{5pt}
    \small
    \begin{tabular}{lccc}
        \toprule
        Tested detector
        & \multicolumn{3}{c}{Source detector used for optimization} \\
        \cmidrule(lr){2-4}
        & F-RCNN & M-RCNN & SSD \\
        \midrule
        F-RCNN & 2.819 & 2.841 & 1.054 \\
        M-RCNN & 2.111 & 2.417 & 0.765 \\
        SSD    & 2.193 & 2.232 & 2.349 \\
        YOLOv3 & 6.098 & 6.180 & 5.199 \\
        YOLOv5 & 4.684 & 4.852 & 3.619 \\
        YOLOv8 & 4.429 & 4.418 & 3.210 \\
        \bottomrule
    \end{tabular}
\end{table}

\begin{table}[t]
    \centering
    \caption{Cross-detector transferability comparison with strong physical attack baselines. All adversarial objects are optimized on YOLOv3 and evaluated on unseen detectors. We report LCR ($\uparrow$), where higher values indicate stronger transferability.}
    \label{tab:baseline_transferability}
    \setlength{\tabcolsep}{4.5pt}
    \small
    \begin{tabular}{lccccc}
        \toprule
        Tested detector
        & \multicolumn{5}{c}{Attack method optimized on YOLOv3} \\
        \cmidrule(lr){2-6}
               & FCA & TT3D & RAUCA & PACG & \textbf{3DGAA} \\
        \midrule
        F-RCNN & 0.4514 & 0.4550 & 0.5086 & 0.3660 & \best{0.6151} \\
        M-RCNN & 0.5827 & 0.5345 & 0.6575 & 0.4390 & \best{0.7948} \\
        SSD    & 1.0483 & 0.9005 & 1.3333 & 0.8302 & \best{1.6386} \\
        YOLOv5 & 0.7939 & 0.8540 & 0.8302 & 0.4841 & \best{1.3061} \\
        YOLOv8 & 0.6812 & 0.7207 & 0.7169 & 0.3032 & \best{0.9348} \\
        \midrule
        Avg.   & 0.7115 & 0.6929 & 0.8093 & 0.4845 & \best{1.0579} \\
        \bottomrule
    \end{tabular}
\end{table}

\vspace{1mm}
\noindent\textbf{Transferability Compared with Physical Attack Baselines.}
Table~\ref{tab:baseline_transferability} addresses the comparison beyond 3DGAA itself by testing representative physical baselines under the same source-target setting. Across all five unseen detectors, 3DGAA obtains the highest LCR, with the largest gains on SSD and YOLOv5. Table~\ref{tab:cross-detector} evaluates source-detector dependence within 3DGAA, whereas Table~\ref{tab:baseline_transferability} evaluates whether 3DGAA remains more transferable than prior physical attacks under a common source detector.

\vspace{1mm}
\noindent\textbf{Segmentation Attack.}
We further test 3DGAA's generalization to segmentation models using mIoU on DeepLabV3 and FCN with ResNet backbones. As shown in Table~\ref{tab:seg_miou}, 3DGAA consistently reduces segmentation quality across all four models. FCN-R50 drops from 74.6\% to 43.0\% mIoU, and FCN-R101 drops from 80.1\% to 43.1\%, corresponding to 31.6\% and 37.0\% absolute mIoU decreases. DeepLabV3 remains stronger after attack, but still shows clear degradation, with mIoU drops of 29.5\% on R50 and 24.8\% on R101. These results confirm that the adversarial object affects semantic segmentation performance under a standard segmentation metric, rather than only reducing detector confidence.

% \begin{table}[t]
%     \centering
%     \caption{LCR ($\uparrow$) of segmentation models under 3DGAA attack. 3DGAA achieves strong performance on segmentation models.}\vspace{-2mm}
%     \small
%     \label{tab:seg_robustness}
%     \begin{tabularx}{\columnwidth}{@{}lXXXX@{}}
%          \toprule
%          Method & \mbox{DLv3-R50} & \mbox{DLv3-R101} & \mbox{FCN-R50} & \mbox{FCN-R101} \\
%          \midrule
%          LCR & 0.628 & 0.874 & 0.795 & 0.893 \\
%          \bottomrule
%     \end{tabularx}\vspace{1mm}
%     \small\raggedright DLv3: DeepLabV3; R50/101: ResNet-50/101 backbones.
%     \vspace{-3mm}
% \end{table}

\begin{table}[t]
    \centering
    \caption{mIoU performance of segmentation models under 3DGAA attack. 3DGAA achieves strong performance on segmentation models.}\vspace{-2mm}
    \label{tab:seg_miou}
    \small
    \begin{tabular}{lccc}
        \toprule
        Model & Clean mIoU $\uparrow$ & Adv. mIoU $\downarrow$ & mIoU Drop $\uparrow$ \\
        \midrule
        FCN-R50 & 74.6\% & 43.0\% & 31.6\% \\
        FCN-R101 & 80.1\% & 43.1\% & 37.0\% \\
        DLv3-R50 & 83.5\% & 54.0\% & 29.5\% \\
        DLv3-R101 & 87.2\% & 62.4\% & 24.8\% \\
        \bottomrule
    \end{tabular}\vspace{1mm}
    \small\raggedright DLv3: DeepLabV3; R50/101: ResNet-50/101 backbones.
    \vspace{-3mm}
\end{table}

\vspace{1mm}
\noindent\textbf{Direct Optimization on Segmentation.}
Under the same $12\times 3$ view protocol, we also directly optimize a segmentation model (DeepLabv3-R50) by suppressing the car class over the object mask $\Omega$.
Concretely, we minimize
\begin{equation}
    \mathcal{L}_{\mathrm{seg}} = \frac{1}{|\Omega|}\sum_{x\in\Omega} p_c(x),
\end{equation}
where $p_c(x)$ denotes the predicted probability of the car class at pixel $x$, and use 
$\mathcal{L}=\lambda_{\mathrm{seg}}\mathcal{L}_{\mathrm{seg}}+\lambda_{\mathrm{shape}}\mathcal{L}_{\mathrm{shape}}+\lambda_{\mathrm{phys}}\mathcal{L}_{\mathrm{phys}}$
as the total objective.
Table~\ref{tab:seg_opt_miou} reports the corresponding mIoU results on DeepLabV3-R50. Detection-driven transfer reduces mIoU from 83.5\% to 54.0\%, yielding a 29.5\% absolute mIoU drop. Direct segmentation optimization further reduces mIoU to 37.8\%, increasing the drop to 45.7\%. This confirms that, under the same view protocol, optimizing a segmentation loss provides additional leverage beyond detector-driven transfer, without changing body geometry. It also shows that 3DGAA can natively attack segmentation models within the same render-and-backprop pipeline.

% \begin{table}[t]
% \centering
% \setlength{\tabcolsep}{6pt}
% \caption{LCR ($\uparrow$) on DeepLabv3-R50 when optimizing via detection and segmentation.} \vspace{-2mm}
% \label{tab:seg_opt}
% \begin{tabular}{lccc}
% \toprule
% Method & Vanilla  & Det\,$\rightarrow$\,Seg & Seg \\
% \midrule
% LCR & 0.000 & 0.628 & 1.144 \\
% \bottomrule
% \end{tabular}\vspace{-4mm}
% \end{table}
\begin{table}[t]
    \centering
    \caption{mIoU comparison under detection-driven transfer and direct segmentation optimization.}
    \label{tab:seg_opt_miou}
    \small
    \begin{tabular}{lcc}
        \toprule
        Method & mIoU $\downarrow$ & mIoU Drop $\uparrow$ \\
        \midrule
        Clean & 83.5\% & -- \\
        Det$\rightarrow$Seg & 54.0\% & 29.5\% \\
        Seg & 37.8\% & 45.7\% \\
        \bottomrule
    \end{tabular}
\end{table}

\subsection{Evaluation Under Input Preprocessing Defenses}
\label{sec:defense_evaluation}
To assess whether simple input-space defenses mitigate the rendered adversarial objects, we evaluated 3DGAA under common preprocessing operations before detector inference. The same adversarial rendered images were used, and only the detector input was transformed. Table~\ref{tab:defense_evaluation} reports mAP, ASR, and LCR under each defense setting.

\begin{table}[t]
    \centering
    \caption{Defense evaluation under common input preprocessing operations. All entries are computed on the same rendered test set used for the detection attack evaluation.}
    \label{tab:defense_evaluation}
    \setlength{\tabcolsep}{3.8pt}
    \small
    \begin{tabular}{lccc}
        \toprule
        Defense & mAP $\downarrow$ & ASR $\uparrow$ & LCR $\uparrow$ \\
        \midrule
        None & 7.28\% & 91.39\% & 3.5628 \\
        JPEG compression & 18.01\% & 78.26\% & 2.2921 \\
        Resize & 21.62\% & 75.40\% & 2.0286 \\
        Gaussian blur & 11.73\% & 83.37\% & 2.9107 \\
        Brightness normalization & 12.68\% & 86.49\% & 2.7984 \\
        \bottomrule
    \end{tabular}
\end{table}

Preprocessing defenses partially restore detector performance but do not eliminate the attack. JPEG compression and resizing increase mAP from 7.28\% to 18.01\% and 21.62\%, respectively, but ASR remains 78.26\% and 75.40\%. This indicates that 3DGAA does not rely only on fragile pixel-level or high-frequency artifacts. Its higher-dimensional 3DGS optimization distributes the adversarial effect across geometry-aware and appearance-aware object parameters, so the attack remains effective after compression and resampling.

The last two defenses also support the role of the Physical Augmentation Module. Gaussian blur and brightness normalization correspond to sensing degradation and photometric variation, which are explicitly modeled during optimization. Under these transformations, 3DGAA still achieves 83.37\% and 86.49\% ASR, respectively. This behavior is consistent with the module ablation in Table~\ref{tab:module_ablation}: training with deployment-time noise, blur, and lighting variation prevents the adversarial object from overfitting to a nominal rendered view.

\subsection{Physical Realization}
\label{sec:physical realization}
To assess physical deployability under the non-deformability constraint, we use a controlled miniature-vehicle setup in which all texture variants are applied to the same rigid shell. Textures are obtained from the appearance-only pipeline $\text{3DGAA}^{a}$ (Sec.~\ref{sec:selective dimension optimization}) and transferred to printable matte wraps, so the vehicle geometry remains unchanged. This protocol provides a geometry-preserving, print-and-apply validation path without body modification.

\begin{figure}[t]
    \centering
    \includegraphics[width=0.95\linewidth]{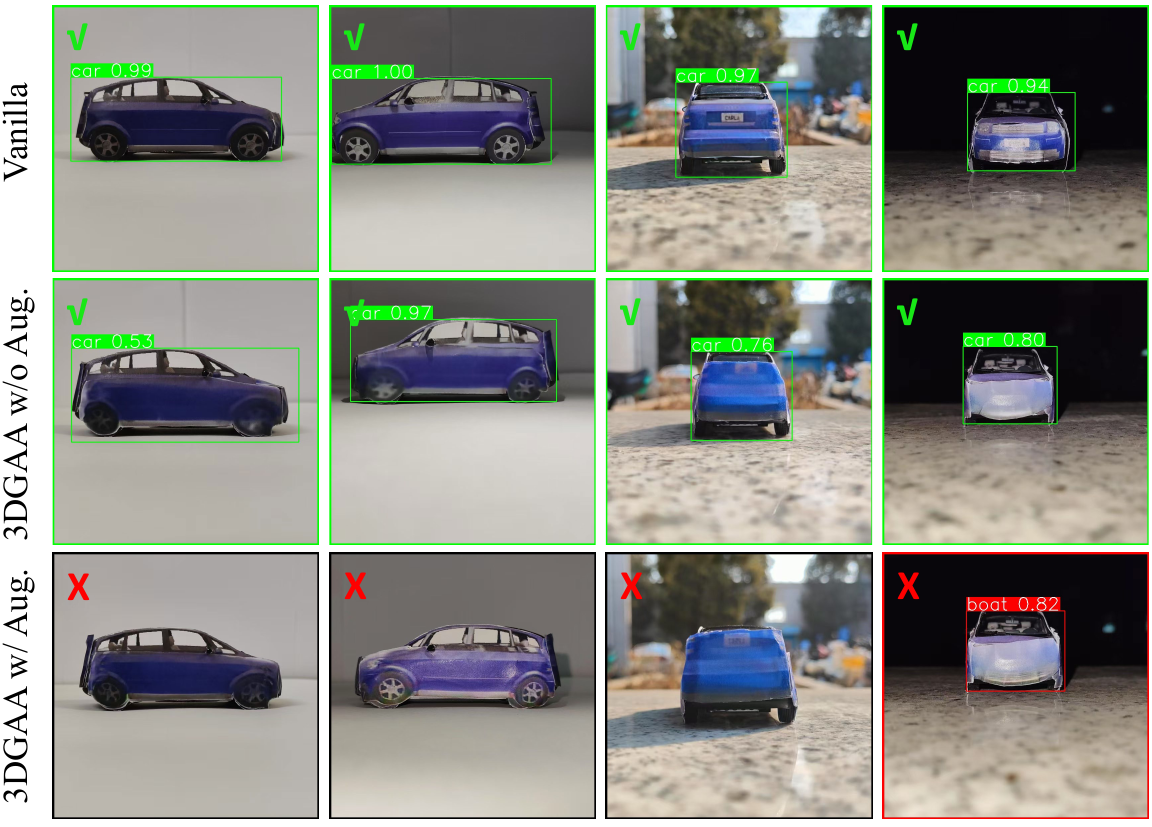}\vspace{-2mm}
    \caption{\textbf{Controlled physical adversarial attack under varying lighting.} Column: ID, IN, OD, ON. Row 1: vanilla, Row 2: w/o Aug., Row 3: w/ Aug. Physical augmentation improves robustness across the tested lighting conditions.}
    \label{fig:physical_realization}\vspace{-2mm}
\end{figure}

We evaluate four physical lighting conditions: indoor daylight (ID), indoor night (IN), outdoor daylight (OD), and outdoor night (ON). Figure~\ref{fig:physical_realization} compares three texture variants: vanilla, 3DGAA without physical augmentation, and 3DGAA with physical augmentation. The vanilla object is reliably detected, whereas the non-augmented adversarial texture is more sensitive to lighting variation. The augmented 3DGAA texture produces substantially weaker detections across the four conditions, which is consistent with the role of physical augmentation in modeling exposure, lighting, and partial occlusion during optimization.

We further re-analyze the available physical records under controlled scale variation. Table~\ref{tab:physical_analysis} reports condition-level mAP and ASR for $\text{3DGAA}^{a}$. The attack remains effective across indoor and outdoor scenes, with an average mAP of 32.08\% and an average ASR of 77.92\%. The night and outdoor daylight conditions show slightly stronger suppression, while indoor daylight and outdoor night remain more challenging. This pattern indicates that the physical effect is not confined to a single favorable scene, but still depends on illumination and apparent object scale.

\revise{Table~\ref{tab:physical comparison} further quantifies detector performance (mAP@0.5) across four architectures under the same miniature-vehicle protocol. With identical geometry and capture protocol, $\textbf{3DGAA}^{a}$ reduces AP from 84.17--96.25\% to 17.92--35.83\%, outperforming the evaluated physical baselines. Together with Sec.~\ref{sec:selective dimension optimization}, these results support a limited but concrete appearance-only fabrication path for vehicle-level physical attacks.}{Table~\ref{tab:physical comparison} reports the matched physical comparison across four architectures. With geometry and capture protocol fixed, $\textbf{3DGAA}^{a}$ obtains 17.92--35.83\% mAP, lower than Vanilla, Random, CAMOU, DAS, and FCA in every column. This complements Table~\ref{tab:physical_analysis} and Fig.~\ref{fig:physical_realization}, which test sensitivity across four lighting settings. Because $\textbf{3DGAA}^{a}$ targets printable wraps, detector performance rather than pattern complexity is the evaluation criterion.}
\revnote{AE/R1: Organized the existing matched baseline comparison and multi-condition physical records as complementary evidence, without adding new experiments.}

\begin{table}[t]
    \centering
    \caption{Condition-level quantitative analysis of controlled physical records under controlled scale variation.}\vspace{-2mm}
    \label{tab:physical_analysis}
    \renewcommand{\arraystretch}{1.2}
    \begin{tabular}{lccccc}
        \toprule
        Metric & ID & IN & OD & ON & Avg. \\
        \midrule
        mAP@0.5 $\downarrow$ & 35.56\% & 28.71\% & 29.07\% & 35.05\% & 32.08\%\\
        ASR $\uparrow$ & 76.67\% & 81.67\% & 80.00\% & 73.33\% & 77.92\% \\
        \bottomrule
    \end{tabular}
\end{table}

\begin{table}[t]
    \centering
    \caption[Matched controlled physical comparison.]{\revise{Controlled physical mAP@0.5 ($\downarrow$) comparison of attack methods under the same miniature-vehicle protocol.}{Matched controlled physical mAP@0.5 ($\downarrow$) comparison under the same miniature-vehicle protocol.}}\vspace{-2mm}
    % \small
    \label{tab:physical comparison}
    \renewcommand{\arraystretch}{1.2}
    \begin{tabularx}{\columnwidth}{@{}>{\hsize=1\hsize}p{1.8cm} 
                                  >{\centering\arraybackslash}p{1cm} 
                                  >{\centering\arraybackslash}p{1cm} 
                                  >{\centering\arraybackslash}p{1.3cm} 
                                  >{\centering\arraybackslash}p{1.3cm}@{}}
         \toprule
         Method & YOLOv5 & SSD & \mbox{F-RCNN} & \mbox{M-RCNN} \\
         \midrule
         Vanilla & 93.75\% & 84.17\% & 96.25\% & 94.58\% \\ 
         Random  & 80.00\% & 76.25\% & 87.50\% & 82.50\% \\
         CAMOU~\cite{Zhang_Foroosh_David_Gong_2018} & 69.58\% & 64.58\% & 72.08\% & 74.17\% \\
         DAS~\cite{Wang_Liu_Yin_Liu_Tang_Liu_2021}   & 75.83\% & 74.17\% & 79.58\% & 77.50\% \\
         FCA~\cite{Wang_Jiang_Sun_Zhou_Gong_Zhang_Yao_Chen_2022} & 60.83\% & 48.33\% & 62.92\% & 67.92\% \\
         $\textbf{3DGAA}^a$ & \textbf{32.08\%} & \textbf{17.92\%} & \textbf{35.83\%} & \textbf{34.58\%} \\
         \bottomrule
    \end{tabularx}\vspace{-5mm}
\end{table}
\revnote{AE/R1: Updated the physical-comparison caption to emphasize the matched protocol.}

%% file: sec/10_conclusion.tex
% \section{Acknowledgments}
% Specific funding information will be included in the camera-ready version upon acceptance.
\vspace{-1mm}
\section{Conclusion}%\vspace{-2mm}
\label{sec:conclusion}

We present \textbf{3DGAA}, a 3D Gaussian-based adversarial object generation framework that jointly optimizes geometry and appearance in the 3DGS parameter space. Unlike prior texture-only methods, 3DGAA enables expressive, multi-view-consistent perturbations while explicitly accounting for physical deployability in safety-critical autonomous driving and related camera-based perception systems. We further introduce a \emph{Physical Filtering Module} that enforces geometric plausibility and a \emph{Physical Augmentation Module} that simulates environmental variations, together yielding adversarial objects that are both realistic and robust under camera-based perception.

Extensive experiments show that 3DGAA significantly degrades detection performance (mAP reduced to 7.38\%) while maintaining physical realism across detectors, viewpoints, and weather conditions. Human-perception studies indicate that the generated textures remain visually plausible, and controlled miniature-vehicle trials provide condition-level physical evidence under representative lighting. Additional evaluations show transfer across viewpoints and model architectures, degradation of segmentation models, and partial resilience to common preprocessing defenses. Together, these results support 3DGAA as a bounded but practical pipeline for adversarial robustness and security evaluation of camera-based perception stacks in connected and autonomous vehicle scenarios.

% A discussion of security concerns, potential societal risks, and mitigation strategies is provided in Sec.~\ref{sec:ethics}.

% A discussion of security concerns and potential societal risks and mitigation strategies is included in App.~\ref{sec:ethics}.

% We present \textbf{3DGAA}, a novel adversarial object generation framework that jointly optimizes geometry and appearance in the 3D Gaussian space. To meet real-car non-deformability, we expose a \emph{dimension-selection} switch, including an \emph{appearance-only} variant (\(\text{3DGAA}^{a}\)) that freezes geometry and supports a geometry-preserving, print-and-apply deployment path. Unlike prior methods, 3DGAA enables expressive, multi-view-consistent adversarial robustness. Furthermore, we propose a \textit{Physical Filtering Module} for physical realism and a \textit{Physical Augmentation Module} for simulating environmental factors to enable real-world deployment.
% Extensive experiments show that 3DGAA significantly degrades detection performance (mAP $\downarrow$ to 7.38\%) while maintaining strong physical realism across detectors and environments. It also generalizes across viewpoints, model architectures, and segmentation tasks, validating 3DGAA as a robust and transferable pipeline for evaluating safety-critical perception systems, especially in autonomous driving scenarios.

% A detailed discussion of security concerns and potential societal risks and mitigation strategies is included in Appendix~\ref{sec:ethics}.

%% file: sec/12_appendix.tex
% \section{Appendix Section}
% \label{sec:appendix_section}
% Supplementary material goes here.

% \setcounter{page}{1}

% \setcounter{figure}{0}
% \setcounter{table}{0}   
% \setcounter{section}{0}
% \setcounter{equation}{0}
% \renewcommand{\thesection}{\Alph{section}} % 主标题字母编�?% \renewcommand{\thesubsection}{\Alph{section}.\arabic{subsection}} 

% \renewcommand\thetable{\Alph{section}.\arabic{table}}
% \renewcommand\thefigure{\Alph{section}.\arabic{figure}}
% % \renewcommand\thesection{A\arabic{section}}
% \renewcommand\theequation{\Alph{section}.\arabic{equation}}
% % \renewcommand\thealgocf{A.\arabic{algocf}}

% \renewcommand*{\theHtable}{\thetable}
% \renewcommand*{\theHfigure}{\thefigure}
% \renewcommand*{\theHequation}{\theequation}
% \renewcommand*{\theHsection}{\thesection}
% \renewcommand*{\theHalgocf}{\thealgocf}

% \twocolumn[\vspace{2cm}]

This supplementary material provides additional method details, experimental results, and discussion of several extended questions.

\section{Supplementary Methods}
Note that the selected intermediate physical derivations have been condensed for conciseness.

\subsection{Physical Augmentation Module}
\label{sup: physical augmentation module}

This section details the exact formulation of the differentiable augmentation components defined in Section~\ref{sec:physical augmentation module}. The physical augmentation module applies a sequence of four transformations:
\begin{equation}
    \mathcal{T} = \mathcal{T}_{\text{occl}} \circ \mathcal{T}_{\text{shadow}} \circ \mathcal{T}_{\text{photo}} \circ \mathcal{T}_{\text{noise}},
\end{equation}
where each sub-transformation models a common source of physical variation. The operations are applied in right-to-left order during optimization.

\textbf{Imaging Degradation} simulates camera sensor noise, which increases with object distance due to atmospheric interference. This degradation is modeled as an additive Gaussian noise function:

\begin{equation}
    \mathcal{T}_{noise} = I + \mathcal{N}(0, \sigma^2(d)),
\end{equation}
where $I$ is the original rendered image, $\mathcal{N}(0, \sigma^2)$ represents zero-mean Gaussian noise, and $\sigma(d)$ is the depth-dependent standard deviation:

\begin{equation}
    \sigma(d) = \sigma_0 + \gamma \cdot d.
\end{equation}
Here, $\sigma_0$ is the base noise level, $d$ represents the pixel-wise depth value, and $\gamma$ is a scaling factor controlling noise amplification with distance.

\textbf{Photometric Variation} simulates color distortions caused by varying lighting conditions and sensor imperfections. We model this effect using channel-wise affine transformations:
\begin{equation}
    \mathcal{T}_{photo} = I_{c} \cdot \alpha_c + \beta_c,
\end{equation}

where $I_{c}$ is the original intensity of color channel $c$, $\alpha_c \sim U(0.9, 1.1)$ represents contrast variation sampled from a uniform distribution, and $\beta_c \sim U(-0.05, 0.05)$ is an additive shift factor.

These transformations ensure that adversarial perturbations remain effective under different lighting conditions.

\textbf{Shadow Projection} used sampling random light source positions and computing penumbra-umbra transitions via sigmoid intensity mapping, we replicated the soft shadow boundaries observed in natural illumination environments.

\begin{equation}
    \mathcal{T}_{shadow} = \frac{1}{1 + e^{-\alpha (d(x, y) - d_{\text{th}})}},
\end{equation}
where $\alpha$ controls shadow smoothness, $d(x, y)$ is the depth at pixel $(x, y)$, and $d_{\text{th}}$ is the threshold depth for shadow casting.

\textbf{Adaptive Occlusion} enhances robustness to partial obstructions through random rectangular masks that simulate large-scale object occlusions covering part of the image area.
\begin{equation}
    \mathcal{T}_{occl} = 
    \begin{cases} 
    \{0, random(0, 1)\}, & (x, y) \in \text{occlusion region}, \\
    1, & \text{otherwise}.
    \end{cases}
\end{equation}

The composite transformation $\mathcal{T} = \mathcal{T}_{occl} \circ \mathcal{T}_{shadow} \circ \mathcal{T}_{photo} \circ \mathcal{T}_{noise}$ defines the Physical Augmentation Module during optimization. Because these transformations are applied before the detector loss is computed, the adversarial object is optimized under the same nuisance factors that later appear in the robustness tests in Section~\ref{sec:exp}.

\subsection{Selective Dimension Optimization}
\label{sup: selective dimension optimization m}

We define three fundamental optimization modes based on parameter space decomposition in the 3D Gaussian Splatting representation:

\textbf{Geometry-only Mode} ($\mathcal{K}_g$): 
This mode activates all 10 geometry-related dimensions, including position ($x$, $y$, $z$), scale ($s_x$, $s_y$, $s_z$), and quaternion rotation ($q_w$, $q_x$, $q_y$, $q_z$):
\begin{equation}
\mathcal{K}_g = \{x, y, z, s_x, s_y, s_z, q_w, q_x, q_y, q_z\}
\end{equation}
This mode enables structural manipulation of the object without altering appearance, suitable for settings like geometric camouflage or shape-aware perturbations.

\textbf{Appearance-only Mode} ($\mathcal{K}_a$): 
This mode only optimizes the 4 appearance-related parameters—color ($c_r$, $c_g$, $c_b$) and opacity ($\alpha$):
\begin{equation}
\mathcal{K}_a = \{c_r, c_g, c_b, \alpha\}
\end{equation}
It is ideal for use cases such as 3D-printed objects or painted surfaces, where modifying geometry is impractical or costly.

\textbf{Full-Dimensional Mode} ($\mathcal{K}_{full}$): 
For completeness, we denote the full parameter set as:
\begin{equation}
\mathcal{K}_{full} = \mathcal{K}_g \cup \mathcal{K}_a
\end{equation}
which enables unconstrained optimization across the entire 14D space.

\vspace{1em}
\textbf{Optimization Behavior.}
As demonstrated in Section~\ref{sec:selective dimension optimization}, different modes exhibit different convergence behavior. Geometry-mode reaches high LCR within 30 epochs, while texture-mode requires longer training (up to 200 epochs) but eventually achieves comparable performance. This difference stems from the sensitivity of geometry dimensions (e.g., $x$, $s$) to detector outputs, as discussed in our LCR sensitivity analysis (Fig.~\ref{fig:dim_analysis}).

\vspace{0.5em}
\textbf{Practical Deployment Implications.}
In scenarios where 3D shape alteration is restricted (e.g., rigid vehicle body), $\mathcal{K}_a$ can be adopted to generate adversarial decals or coatings. For applications involving laser-cut, foam, or parametric shell objects, $\mathcal{K}_g$ can produce printable adversarial shapes while maintaining uniform color.

% Further visual examples under different dimension selection strategies are presented in Appendix~\ref{sup: selective dimension optimization}.

% \section{Supplementary of Experiments}
% \begin{figure*}
%     \centering
%     \includegraphics[width=0.6\textwidth]{figs/3DGAA-selective-dim-1x3.pdf}
%     \caption{Selective Dimensions. Mode-selected dimension optimization visualization.}
%     \label{fig:selective 1x3}
% \end{figure*}

% \begin{figure*}
%     \centering
%     \includegraphics[width=\textwidth]{figs/3DGAA-selective-dim-1x6.pdf}
%     \caption{Selective Dimensions. Single dimension optimization visualization.}
%     \label{fig:selective 1x6}
% \end{figure*}

% \begin{figure*}
%     \centering
%     \includegraphics[width=\textwidth]{figs/object_generalization.pdf}
%     \caption{Adversarial Effectiveness of 3DGAA Across Different Vehicle Types.}
%     \label{fig:object generalization}
% \end{figure*}

% \section{Supplementary of Experiments}

\subsection{Implementation Details}
% \section{Implementation Details}
\label{app:implementation details}
The 3DGS generation backbone adopts the pretrained LGM architecture \cite{tang2024lgmlargemultiviewgaussian} with frozen parameters, processing four calibrated views ($512\times512$ resolution) as input. The physical filtering module applies density-based pruning at $\tau_d=0.105$ percentile threshold, followed by structural Gaussian smoothing, as detailed in Sec.~\ref{sec:filtering}.

For adversarial optimization, we initialize the learning rate $\eta=0.03$ with gradient descent over 50 epochs, optimizing all 14 parameters of the 3D Gaussians (positions $x\in\mathbb{R}^3$, rotations $q\in\mathbb{R}^4$, scales $s\in\mathbb{R}^3$, colors $c\in\mathbb{R}^3$, opacity $\alpha\in\mathbb{R}^1$) under the combined loss from Eq.~\ref{equ:total_loss}. The physical augmentation module probabilistically applies transformations with $\gamma=0.005$ for scaling factor and $p_{size}=0.1$ for occlusion size.

All experiments are conducted on an NVIDIA RTX 4090 GPU with PyTorch 2.1 + CUDA 11.8, where each adversarial optimization completes within seconds-to-minutes as reported in Table~\ref{tab:efficiency}.

\subsection{Multi-Angle Visualization of 3DGAA Results}
\begin{figure}
    \centering
    \includegraphics[width=0.95\linewidth]{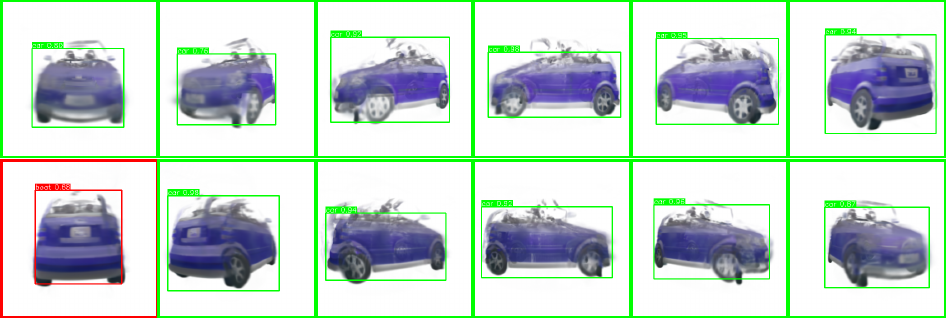}
    \caption{Multi-view visualization of original 3DGS reconstructions: twelve representative viewpoints demonstrating baseline geometric accuracy and texture fidelity.}
    \label{fig:sup_orig}
\end{figure}

\begin{figure}
    \centering
    \includegraphics[width=0.95\linewidth]{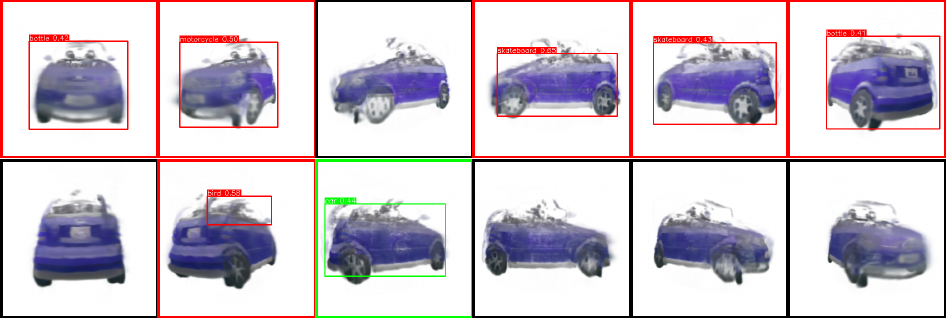}
    \caption{Consistent adversarial effects across viewing angles: corresponding adversarial 3DGS visualizations under identical viewpoints, showing maintained structural integrity while evading detection.}
    \label{fig:sup_adv}
\end{figure}

Figures~\ref{fig:sup_orig} and~\ref{fig:sup_adv} provide extended multi-perspective visualizations of our method's capability to maintain structural authenticity while achieving adversarial effectiveness. Figure~\ref{fig:sup_orig} demonstrates the original 3DGS reconstructions across 12 representative viewpoints, highlighting the baseline geometric accuracy and texture fidelity. Correspondingly, Figure~\ref{fig:sup_adv} showcases the adversarial counterparts under identical viewing conditions, where optimized texture patterns consistently mislead detectors without introducing noticeable visual disparity.

The side-by-side comparison reveals two key observations: (1) The adversarial textures preserve high-frequency surface details comparable to original patterns, ensuring physical plausibility, and (2) adversarial effectiveness remains stable across extreme viewing angles, confirming the 3DGAA optimization's robustness to perspective variations. These visual results complement our quantitative analyses by demonstrating the spatial consistency of geometric reconstruction and adversarial pattern generation.

\section{Discussion}

\subsection{Adversarial Effectiveness is Derived from 3DGAA}
The adversarial effectiveness of 3DGAA originates from its explicit adversarial optimization, which differentiates it from conventional 3D Gaussian Splatting (3DGS) representations. While 3DGS naturally introduces slight distortions in shape and texture due to the limitations of generation and sampling, these variations do not inherently contribute to adversarial behavior. The core objective of 3DGS is to reconstruct high-fidelity 3D objects from multi-view images, ensuring that the generated representation maintains visual consistency rather than misleading object detection models.

In contrast, 3DGAA explicitly optimizes the adversarial properties of the object by introducing targeted perturbations in texture and geometry. This process is guided by an adversarial loss that reduces detection confidence, while a shape-preservation constraint limits excessive geometric distortion that would otherwise compromise physical plausibility.

Our sensitivity analysis in Sec.~\ref{sec:exp_selective} further clarifies where this adversarial power comes from: geometric parameters ($x,s$) drive rapid early-stage LCR gains, while appearance parameters ($\alpha,c$) enable stronger long-term degradation once given more iterations. Rotation $q$ is comparatively insensitive. This pattern explains why full 3DGAA (optimizing both geometry and appearance) achieves the strongest overall effect, while $\text{3DGAA}^{a}$ remains attractive as a geometry-preserving, appearance-only realization for physical deployment.

\subsection{Discussion on SSD Transferability}
The SSD results are best interpreted as detector-dependent transferability. SSD is a one-stage dense detector whose decisions depend on default-box matching and scale-localized feature responses, whereas two-stage detectors use region proposals followed by classification. A transferred perturbation may therefore reduce global confidence without equally disrupting SSD's anchor-level evidence. This does not indicate intrinsic robustness to 3DGAA: when SSD is used as the source detector, Table~\ref{tab:cross-detector} remains strong, and under the shared YOLOv3-to-unseen-detector protocol 3DGAA still gives the highest SSD LCR among the physical attack baselines in Table~\ref{tab:baseline_transferability}.

\subsection{Generalization to Non-Vehicle Objects}
Although our experiments focus on vehicles, 3DGAA optimizes object-level 3DGS parameters rather than vehicle-specific semantics. This suggests possible extension to other rigid objects after category-specific capture and fabrication constraints are checked.

Within vehicles, the same protocol applies across different body types, while the appearance-only mode remains the most direct geometry-preserving route for physical adaptation.

Broader deployment to non-vehicle objects remains future work because each category may impose different geometry, material, and capture constraints.

\subsection{Ethical Considerations}
\label{sec:ethics}

3DGAA is intended for controlled robustness evaluation of camera-based perception systems, but realistic adversarial objects could be misused outside research settings. We therefore restrict all attacks to simulation or miniature-scale controlled experiments, do not conduct field deployment, and do not publicly release attack code or physical models at this stage. The method is presented to support vulnerability analysis and future defensive design. The perceptual user study in Sec.~\ref{sec:comprehensive comparison} was anonymous and non-invasive, collected no personally identifiable or biometric information, and only asked participants to rate rendered object realism.

\subsection{Limitations and Future Work}
The present physical study focuses on controlled miniature-scale deployment, where geometry, texture application, lighting condition, and camera capture can be compared under a consistent protocol. This design is appropriate for verifying the geometry-preserving print-and-apply path, but it does not exhaust the full range of dynamic road conditions. Future work will extend this evaluation to moving platforms, stronger motion blur, longer video trajectories, and synchronized frame-level detection analysis. Such tests require a dedicated video-level protocol so that adversarial effects can be separated from ordinary detection failures caused by motion, defocus, or unstable capture. We therefore interpret the physical results as bounded physical evidence, supported by the multi-view simulation, defense, transferability, and module-ablation analyses in the main text.